\documentclass[nohyperref]{article}

\usepackage{microtype}
\usepackage{graphicx}
\usepackage{booktabs} 

\usepackage{hyperref}
\usepackage{caption}
\usepackage{subcaption}

\usepackage[accepted]{icml2022}

\usepackage{amsmath}
\usepackage{amssymb}
\usepackage{mathtools}
\usepackage{amsthm}

\usepackage[capitalize,noabbrev]{cleveref}

\theoremstyle{plain}
\newtheorem{theorem}{Theorem}[section]

\theoremstyle{definition}

\theoremstyle{remark}

\newcommand{\cumh}{\widehat{cu}}
\newcommand{\cums}{\overset{*}{cu}}
\usepackage[textsize=tiny]{todonotes}

\icmltitlerunning{Towards Invariance to Gradient Descent's Hyperparameter Initialization}

\begin{document}

\twocolumn[
\icmltitle{Read the Signs\\Towards Invariance to Gradient Descent's Hyperparameter Initialization}



\icmlsetsymbol{equal}{*}

\begin{icmlauthorlist}
\icmlauthor{Davood Wadi}{yyy}
\icmlauthor{Marc Fredette}{yyy}
\icmlauthor{Sylvain Senecal}{sch}
\end{icmlauthorlist}

\icmlaffiliation{yyy}{Department of Decision Science, HEC Montreal, Montreal, QC, Canada}
\icmlaffiliation{sch}{Department of Marketing, HEC Montreal, Montreal, QC, Canada}

\icmlcorrespondingauthor{Davood Wadi}{davood.wadi@hec.ca}

\icmlkeywords{Machine Learning, Meta Optimization, Gradient Descent, Convex Optimization}

\vskip 0.3in
]



\printAffiliationsAndNotice{}  

\begin{abstract}
We propose ActiveLR, an optimization meta algorithm that localizes the learning rate, $\alpha$, and adapts them at each epoch according to whether the gradient at each epoch changes sign or not. This sign-conscious algorithm is aware of whether from the previous step to the current one the update of each parameter has been too large or too small and adjusts the $\alpha$ accordingly. We implement the Active version (ours) of widely used and recently published gradient descent optimizers, namely SGD with momentum, AdamW, RAdam, and AdaBelief. Our experiments on ImageNet, CIFAR-10, WikiText-103, WikiText-2, and PASCAL VOC using different model architectures, such as ResNet and Transformers, show an increase in generalizability and training set fit, and decrease in training time for the Active variants of the tested optimizers. The results also show robustness of the Active variant of these optimizers to different values of the initial learning rate. Furthermore, the detrimental effects of using large mini-batch sizes are mitigated. ActiveLR, thus, alleviates the need for hyper-parameter search for two of the most commonly tuned hyper-parameters that require heavy time and computational costs to pick. We encourage AI researchers and practitioners to use the Active variant of their optimizer of choice for faster training, better generalizability, and reducing carbon footprint of training deep neural networks.
\end{abstract}

\section{Introduction}
The current state-of-the-art optimization algorithms are mechanical in their nature, 
in the sense that they passively average the gradients, based on a predefined number of past steps, 
instead of actively adjusting the step size at each epoch according to the change in the sign of the gradients. 
The moving-average-based optimizers, such as SGD with momentum \cite{sutskever2013importance}, 
Adam \citep{kingma2014adam}, RAdam \citep{liu2019variance}, and RMSprop \citep{tieleman2012rms}, 
base their optimization calculations on a fixed number of previous gradient values, regardless of what their sign is. 
If the sign of the gradients changes after epoch $e$, we want the optimizer to recognize this jump over 
the optimum immediately and decrease the step size accordingly. Furthermore, current optimizers update 
all the weights by a global scalar learning rate, $\alpha$ (referred to by most deep learning frameworks as LR or learning rate).

We introduce ActiveLR, an optimization meta algorithm that can be implemented on top of current optimizers 
to adjust the hyper-parameter $\alpha$ at each epoch for each model parameter. We mathematically prove that 
ActiveLR variant of a given optimizer has a lower objective function cost compared to the vanilla implementation of that optimizer. 
One of the main objectives of ActiveLR is to obviate the need for the tuning of the initial learning rate, $\alpha$, 
and mini-batch size. For different datasets, architectures, and tasks, the values of $\alpha^*$, the optimal 
initial learning rate, and the optimal mini-batch size \citep{Iiduka2021TheNO}, which lead to optimal convergence, are different. 
Although for well-known benchmark tasks, such as ImageNet with ResNet18 for object recognition, 
the optimal values for different hyper-parameters have been immensely studied and made publicly available, for new tasks, 
researchers and practitioners must start anew and perform costly hyper-parameters search to find the optimal values 
for the initial learning rate, mini-batch size, what learning rate scheduler to use, and at which epochs they should 
decay the learning rate by what amount. 

Moreover, learning rate values higher than $\alpha^*$ lead to divergence of the optimization, while values lower than $\alpha^*$ 
lead to significantly slower rate of convergence and also increase the possibility of being stuck in a local minimum. 
Also, large mini-batch sizes are necessary for fast training of large-scale datasets and utilizing the full 
computational power of multiple GPUs \citep{You2017ScalingSB}. Our tests show that vanilla implementations of SGD with momentum, 
AdamW, RAdam, and Adabelief get stuck in local minima for smaller learning rates and fail to achieve high performance on the test set. 
Moreover, their performance has a negative correlation with mini-batch size. As we increase the mini-batch size, 
the training becomes unstable and generalizability suffers significantly. On the other hand, 
ActiveLR implementations of these optimizers (i.e., ActiveSGD, ActiveAdamW, ActiveRAdam, and ActiveBelief) 
outperform their original implementations, and are also robust to the values of initial learning rate and mini-batch size. 

This has major implication for research and practice. Not only is tuning the learning rate and mini-batch size 
time consuming and costly, it causes severe environmental impacts. For instance, the greenhouse gas emissions 
for training a state-of-the-art transformers model is equivalent to 10 years of emissions of a person in the U.S. 
\citep{lacoste2019quantifying, strubell2019energy}. Furthermore, while researchers at large institutions have access 
to high compute power and afford to perform intense hyper-parameter search on hundreds of GPUs and TPUs, most AI researchers 
and practitioners have significantly more limited access to computation power. Therefore, ActiveLR helps in democratizing AI, 
saving time and cost to AI researchers, and also reducing greenhouse emissions.

\section{Problematization}

SGD and adaptive optimizers have two shortcomings. First, when the optimizer approaches the optimum, the accumulated momentum 
will cause it to oscillate over the optimum. It is because when the gradient changes its sign, if the exponential moving average 
has the opposite sign, the weight of the parameter will be updated up the slope of the error curve, deviating the model from 
the optimum. Therefore, it takes longer for such optimizers to stabilize around the optimum compared to an optimizer that 
is aware of the gradients’ sign at each step and adjusts the step size, $\alpha$, according to whether the gradients have changed 
their sign or not.\\
Second, using a global learning rate, $\alpha$, to update all the parameters does not account for the specific 
position of each neuron compared with its optimum value. At each training epoch, some neurons are farther away 
from their optimum---requiring a larger $\alpha$---while others are closer to their optimum---requiring a smaller $\alpha$. 
In addition, for a deep neural network, the gradient for middle layers are much smaller compared to the initial 
and final layers (see \ref{gradNorms}), requiring a larger $\alpha$.
Moreover, when we simultaneously update the weights of a previous layer (e.g. $layer_{i-1}$) to correct the same error function, 
this simultaneous change in the incoming weights to the next layer ($layer_{i}$) causes an overshoot effect—determined 
by the term “fan-in”—on the weights of $layer_{i}$. The size of fan-in determines the amount of input a layer receives 
and varies from layer to layer \citep{tieleman2012lecture}. Therefore, we need local $\alpha$'s for each neuron to account 
for these disparities between different neurons.

\section{Related work}
\begin{figure}[ht]
   \vskip 0.2in
   \begin{center}
   \centerline{\includegraphics[width=\columnwidth]{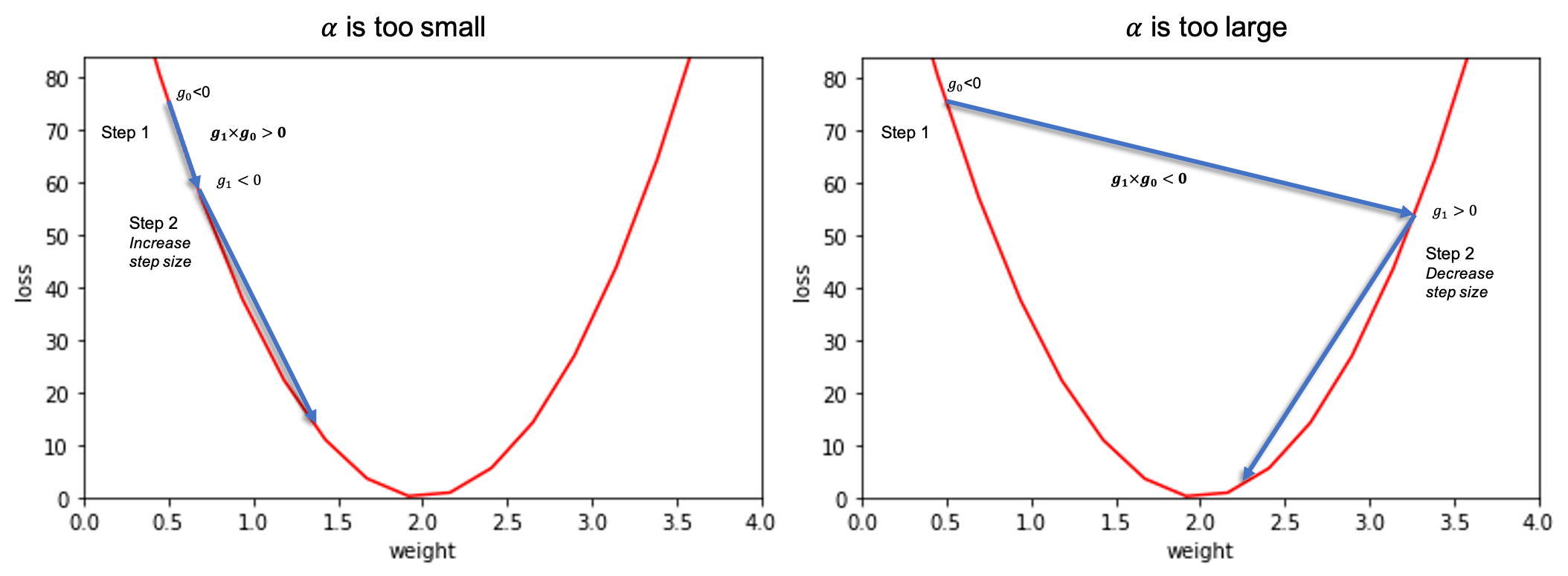}}
   \caption{Optimization of a parameter of a model using a sign-conscious optimizer. When the sign of the gradient does not change (\emph{left}), the Active optimizer increases the step size $\alpha$, nudging the model weight to the true value of the weight ($2.0$). When the sign of the gradient changes (\emph{right}), the weight has jumped over the optimum. Therefore, the Active optimizer decreases the step size $\alpha$ for the next iteration. This not only decreases convergence time, but also avoids constantly jumping over the optimum value.}
   \label{illustration}
   \end{center}
   \vskip -0.2in
\end{figure}

\citet{jacobs1988increased} was the first to suggest that every weight of a neural network be given its own $\alpha$, 
and each $\alpha$ be allowed to vary over time. He states that when the sign of the derivative for a parameter is constant 
for consecutive epochs, the $\alpha$ for that parameter should be increased. On the other hand, when the sign of the derivative 
for a parameter changes for consecutive epochs, the $\alpha$ for that parameter should be decreased (Figure \ref{illustration}). 

Efforts have been made to reduce the sensitivity of gradient descent optimization to learning rate initialization 
\cite{baydin2017online, zhang2019lookahead}. However, previous work on reducing sensitivity to initial learning rate 
has introduced additional hyper-parameters that need to be tuned, which seems to defeat the purpose--removing a hyper-parameter 
and introducing others to be tuned.

To best of our knowledge, there has not been an automatic algorithm that reduces sensitivity to both initial learning rate 
and mini-batch size.

The latest mention of a sign-conscious optimization algorithm was in \cite{tieleman2012lecture}. 
However, since previous implementations worked only on full-batch training and researchers almost never 
want to train their neural networks with full batches, the development of sign-conscious optimizers was thwarted.

\section{ActiveLR}
We introduce a meta algorithm that utilizes a vanilla optimization algorithm, such as Adam or SGD, in the inner-loop part, 
and in the outer-loop part of the optimization, at every epoch, actively adjusts the local learning rate for each parameter 
based on the change in the sign of the cumulative gradients. Unlike previous implementations, ActiveLR works with mini-batch 
and stochastic gradient descent. Almost always, mini-batch gradient descent is the optimal way to train neural networks. 
Firstly, modern datasets (e.g., ImageNet) do not fit into GPU memory, so full-batch gradient descent on such datasets is not possible. 
Furthermore, researchers prefer to use mini-batches instead of full-batches, as mini-batches provide faster convergence 
and better generalizability \citep{qian2020impact, wilson2003general}.

\subsection{The ActiveLR meta algorithm}
Assuming we have $k$ mini-batches of data in our dataset, for each iteration t in the dataset, 
given that each mini-batch of training data has a specific gradient w.r.t. the parameter 
$\theta_i$, $\nabla f^t (\theta_i^{t-1})$, that is calculated for that mini-batch, 
we define the cumulative gradient of the model as the arithmetic summation of all mini-batch gradients. 
For epoch $e$, the cumulative gradient, $\nabla f_{cu}^e$, with respect to the inner-loop-updated parameter, 
$\theta_i$, can be derived from:
\[\nabla f_{cu}^e (\theta_i)=\sum_{t=1}^k \nabla f^t (\theta_i^{t-1})\]
The cumulative gradient $\nabla f_{cu}^e(\theta_i)$ is the gradient of the objective function $f$ with respect to the parameter $\theta_i$ at epoch $e$, as if the model has experienced the whole dataset (full-batch training). Since after seeing each mini-batch the optimizer updates the parameters, the calculated gradients for each mini-batch is different from the gradient of the mini-batch if there was no update (this is what we are trying to approximate). Consequently, we wish to prove that the sign of the gradient for the whole dataset with respect to parameter $\theta_i$ when the parameter does not change inside the loop is the same as that of the updated parameter within the loop. Thus, we prove the Theorem 1 that enables us to use gradient-descent-updated gradients instead of no-update gradients.

\begin{theorem} \label{theorem:1}
In the local convex regime of a non-convex objective function, the cumulative gradient of parameter $\theta_i$ at epoch $e$ with no inner-loop updates, $\cums$, has the same sign as the cumulative gradient of parameter $\theta_i$ at epoch $e$ with inner-loop updates, $\cumh$, if the learning rate, $\alpha$, is smaller than $\alpha^*$ that causes the inner-loop to diverge.
\[ \cums \cumh \geq 0\]
\end{theorem}
ActiveLR can now be implemented using mini-batches by comparing the sign of the cumulative gradient at epoch 
$\mathit{e-1}$, $\nabla f_{cu}^{e-1}(\theta_i)$, with the sign of the cumulative gradient at epoch $e$, 
$\nabla f_{cu}^e(\theta_i)$ (Algorithm \ref{algo2}). 
It allows the optimizer to update the parameters at each step of the mini-batch \citep{wilson2003general, qian2020impact}, 
while adjusting the $\alpha$ at the end of each epoch \citep{tieleman2012lecture}, enabling inner-loop learning 
through the backend optimizer (e.g., Adam, SGD) and active learning rate adjustment together.

\begin{algorithm}[tb]
 \caption{ActiveLR for SGD (ActiveSGD)}
 \label{algo2}
\begin{algorithmic}
  \STATE {\bfseries Inputs:} $\alpha^0$ \COMMENT{initial learning rate}, $\theta_i^0$ \COMMENT{initial parameter $i$}, $f(\theta)$ \COMMENT{stochastic objective function},
$\alpha_{high}$ \COMMENT{$\alpha$ growth constant}, $\alpha_{low}$ \COMMENT{$\alpha$ shrink constant}
  \STATE {\bfseries Output:} $\theta_i^T$ \COMMENT{the updated parameter $i$}
  \STATE $e \gets 0$, $g_{i,cu}^0 \gets 0$ \COMMENT{initialization}
  \REPEAT
  \STATE $e \gets e+1$ \COMMENT{next epoch}
  \STATE $t \gets 0$
  \STATE $g_{i,cu}^e \gets 0$ \COMMENT{set/reset the cumulative gradient to zero at each epoch}
  \FOR{mini-batches in dataset}
  \STATE $t \gets t+1$	
  \STATE $g_i^t \gets \nabla_{\theta_i} f^t(\theta_i^{t - 1})$ \COMMENT{calculate the gradient of the objective w.r.t. the parameter at timestep $t$}
  \STATE $g_{i,cu}^e \gets g_{i,cu}^e + g_i^t$ \COMMENT{add mini-batch gradient to cumulative gradient}
  \STATE $\theta_i^t \gets \theta_i^{t-1} - \alpha_i^t g_i^t$ \COMMENT{SGD update}
  \ENDFOR 
  \STATE $\alpha_i^e \gets
	\begin{cases}
	\alpha_i^{e-1} + \alpha_{high},&  \text{\small $sign(g_{i,cu}^e \times g_{i,cu}^{e-1}) > 0$}\\
	\alpha_i^{e-1} \times \alpha_{low},&  \text{\small $sign(g_{i,cu}^e \times g_{i,cu}^{e-1}) \leq 0$}
	\end{cases}
  $ \COMMENT{adjust learning rate}
  	\UNTIL{$\theta_i^t$ converged}
  \STATE {\bfseries Return:} $\theta_i^T$
\end{algorithmic}
\end{algorithm}

\subsection{Orthogonality of ActiveLR to other optimizers}
A major advantage of ActiveLR is the orthogonality of the hyper-parameter that it adjusts, $\alpha$, 
to the hyper-parameters that other well-known algorithms adjust. To be more specific, we take a look 
at the generic parameter update that is the backbone of SGD, Adam, RAdam, etc.
\begin{equation}\label{eq3}
\theta_i^t  \gets \theta_i^{t-1}-\alpha F(g_i^t)
\end{equation}

The difference among these algorithms is the way they manipulate the $F(g_i^t)$ term through the $F$ function, 
while keeping $\alpha$ as is in the original SGD algorithm. For instance, in the case of Adam, 
$F(g_i^t)$ is $\frac{\frac{\beta_1 m_i^{t-1} + (1-\beta_1) g_i^t}{1-\beta_1^t}}{\sqrt{\frac{\beta_2 v_i^{t-1} + (1-\beta_2) {g_i^t}^2}{1-\beta_2^t}}+\epsilon}$. 
However, since ActiveLR modifies the $\alpha$ term in Equation \ref{eq3}, 
it can be combined with other optimization algorithms. In fact, in our experiments we report results on ActiveLR 
combined with SGD with momentum (ActiveSGD), AdamW (ActiveAdamW), with RAdam (ActiveRAdam), and AdaBelief (ActiveBelief) 
and compare the Active results with their original, vanilla variants.

\section{Convergence analysis}
At any given epoch, $e$, we are faced with the choice between the ActiveLR algorithm vs. 
its corresponding vanilla backbone. We prove that the value of the objective function for ActiveSGD (ActiveLR combined with SGD) 
at epoch $\mathit{e+1}$ is as good or better than the vanilla SGD (backbone) algorithm at epoch $\mathit{e+1}$, 
and show the conditions where each inequality holds. The proof is capable of any first-order backbone algorithm. 
For other backbones, such as Adam, RMSProp, etc., their respective ActiveLR implementation should have the same lines of proof, 
with some modifications peculiar to the optimizer of interest.

\begin{theorem}\label{theorem:2}
Let us call the cost of vanilla SGD at epoch e, $f(\theta^S_e)$, 
and the cost of its corresponding ActiveLR implementation, $f(\theta^A_e)$. 
We define ActiveSGD's gradient, $g_{e+1}^A$, as $\frac{\partial{f(\theta_e^A)}}{\partial{\theta_e^A}}$, 
and vanilla SGD's gradient, $g_{e+1}^S$, as $\frac{\partial{f(\theta_e^S)}}{\partial{\theta_e^S}}$. $\alpha$ 
is the initial learning rate for ActiveSGD and also the constant learning rate for SGD. $\alpha_{High}$ 
is the higher learning rate ($\alpha_{High}>\alpha$) that ActiveLR uses when $g_e^A g_{e+1}^A >0$ and $\alpha_{Low}$ 
the lower learning rate ($\alpha_{Low}<\alpha$) that ActiveLR uses when $g_e^A g_{e+1}^A <0$. 
In the local convex regime of $f$, at any arbitrary epoch, $\mathit{e}$, the difference between 
the cost of using vanilla SGD and ActiveSGD at the next epoch, $\mathit{e+1}$, is

\begin{equation} \label{ineq}
\begin{cases}
f(\theta^S_{e+1}) - f(\theta^A_{e+1}) \geq g_{e+2}^A g_{e+1}^A (\alpha_{High}-\alpha),\\ \text{if } g_e^A g_{e+1}^A >0.\\

f(\theta^S_{e+1}) - f(\theta^A_{e+1}) \geq g_{e+2}^A g_{e+1}^A (\alpha_{Low}-\alpha),\\ \text{if } g_e^A g_{e+1}^A <0.
\end{cases}
\end{equation}
where the right hand side for both cases is non-negative.
\end{theorem}
In other words, ActiveSGD is at least as good as vanilla SGD when gradients are zero (equality in \ref{ineq}), 
and strictly better than vanilla SGD when gradients are not zero (inequality in \ref{ineq}). 
The lower bound of the advantage is $g_{e+2}^A g_{e+1}^A (\alpha_{High}-\alpha)$, 
when $g_e^A g_{e+1}^A >0$ and $g_{e+2}^A g_{e+1}^A (\alpha_{Low}-\alpha)$ when $g_e^A g_{e+1}^A <0$.

\begin{figure}[ht]
    \vskip 0.2in
    \begin{center}
    \begin{subfigure}[b]{0.3\columnwidth}
        \centerline{\includegraphics[width=\columnwidth]{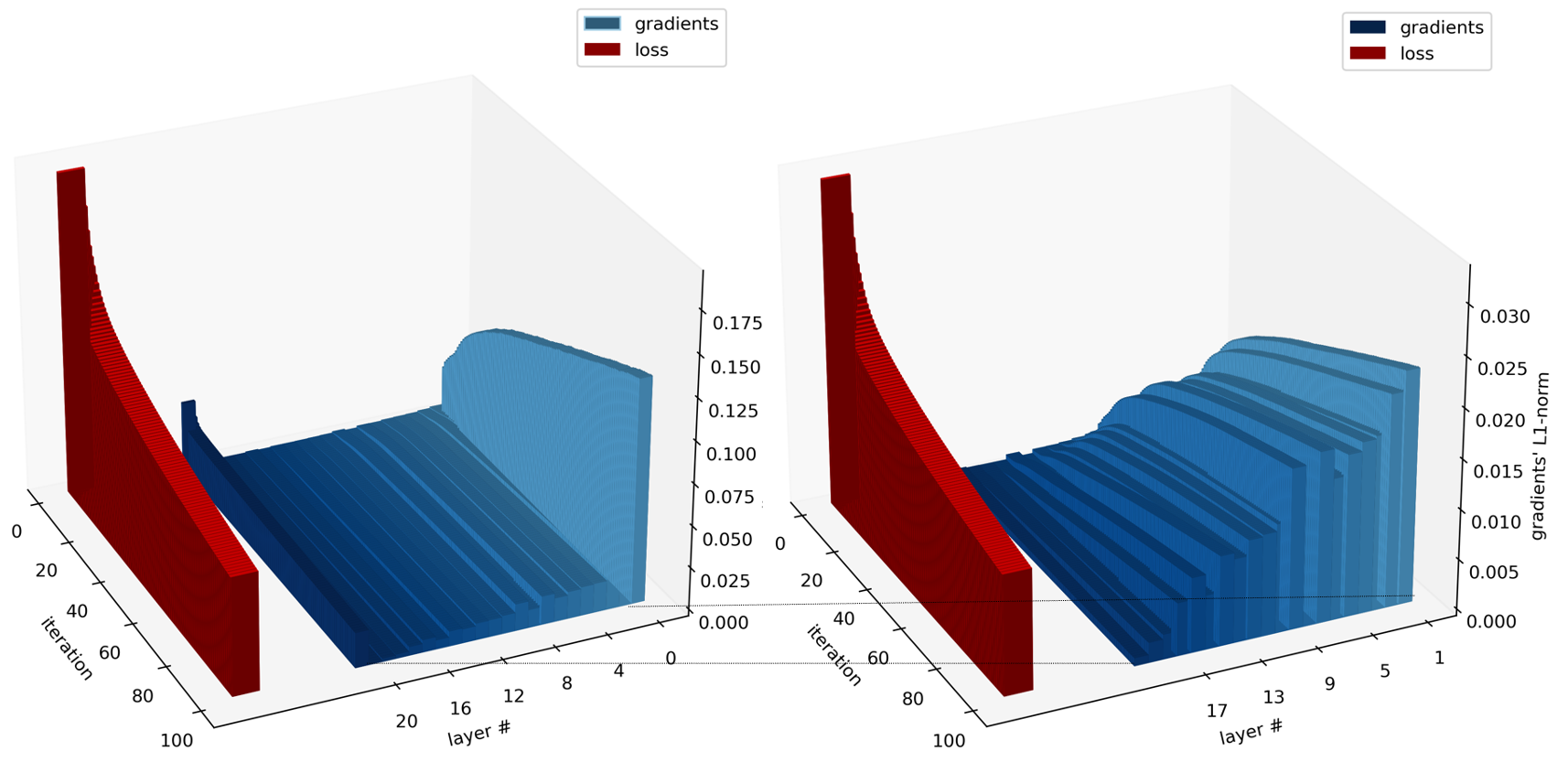}}
        \caption{Adam, $\alpha=10^{-5}$}
        \label{fig:gradLayersA5}
    \end{subfigure}
    \hfill
    \begin{subfigure}[b]{0.3\columnwidth}
        \centerline{\includegraphics[width=\columnwidth]{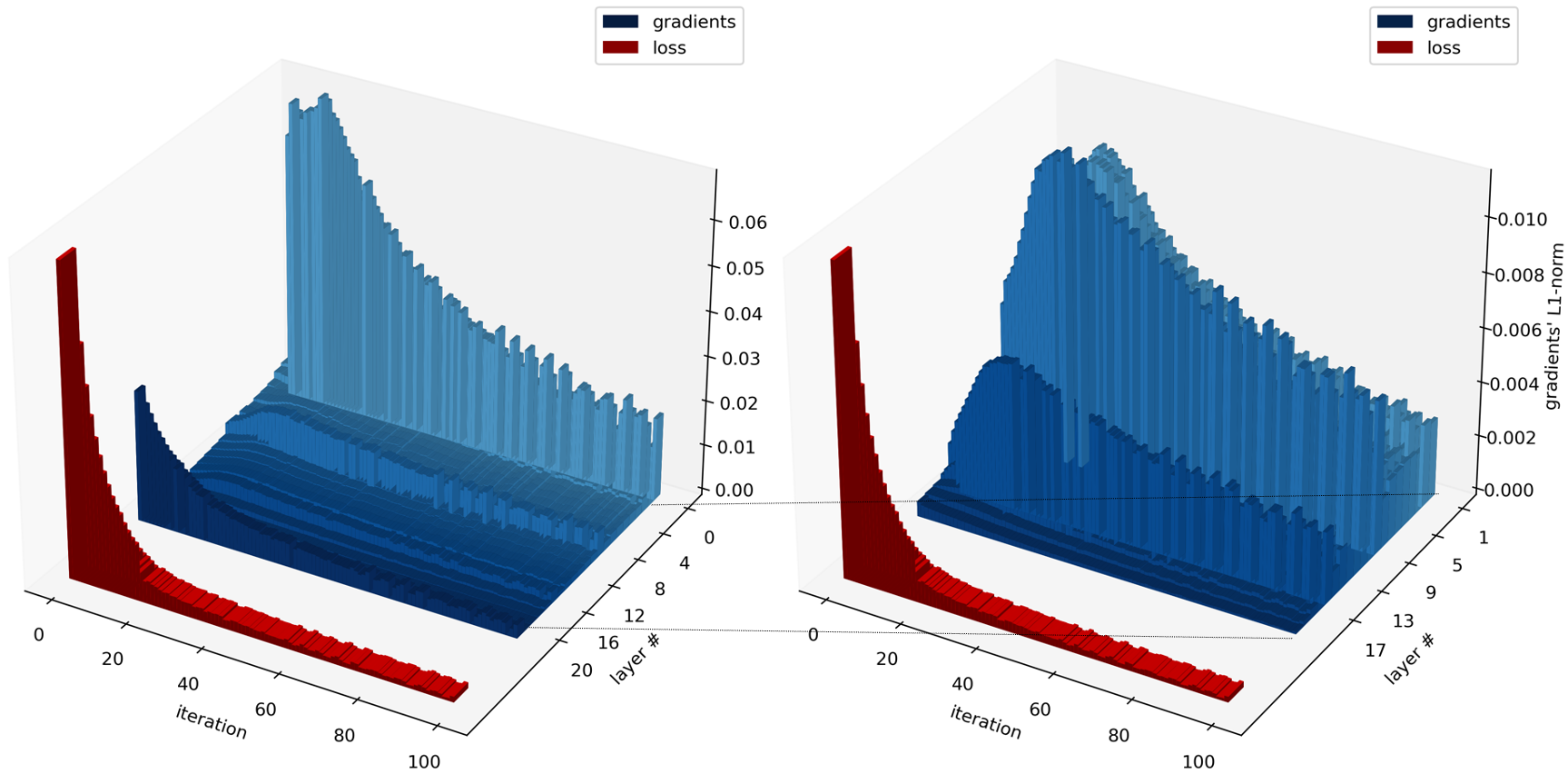}}
      \caption{Adam, $\alpha=10^{-3}$}
      \label{fig:gradLayersA3}
    \end{subfigure}
    \hfill
    \begin{subfigure}[b]{0.3\columnwidth}
        \centerline{\includegraphics[width=\columnwidth]{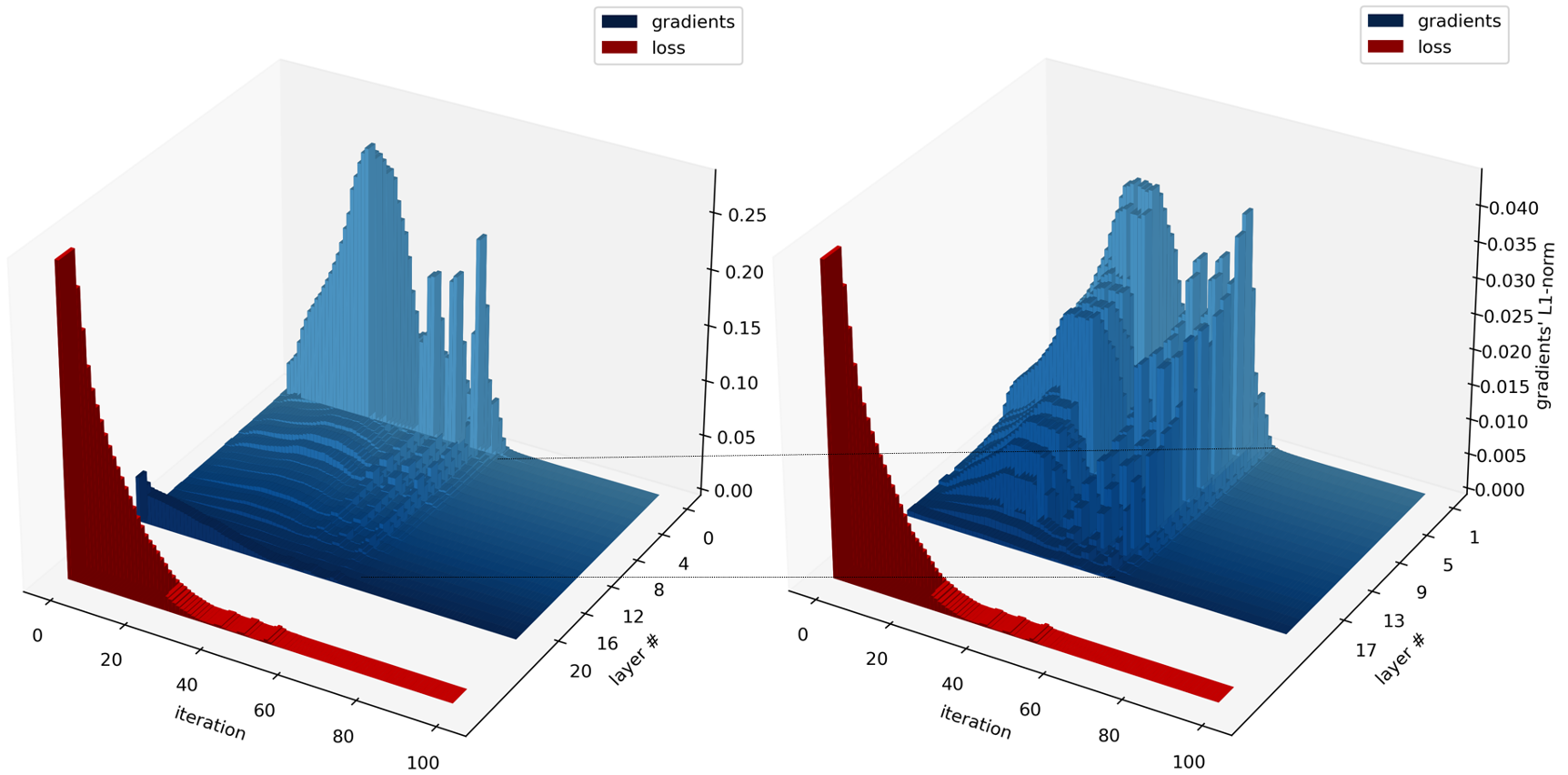}}
      \caption{ActiveAdam, $\alpha=10^{-5}$}
      \label{fig:gradLayersC5}
    \end{subfigure}
    \caption{\textbf{Gradient norms across layers for CIFAR-10 on ResNet-18}. Vanilla Adam when the $\alpha$ is too small ($\alpha=10^{-5}$) gets stuck in a local minimum \subref{fig:gradLayersA5}. Vanilla Adam when the learning rate is optimal ($\alpha=10^{-3}$) oscillates around the optimum \subref{fig:gradLayersA3}. ActiveAdam with the same low learning rate that gets vanilla Adam stuck ($\alpha=10^{-5}$) achieves minimal loss and remains in the optimum \subref{fig:gradLayersC5}.}
   \label{fig:gradLayers}
    \end{center}
   \vskip -0.2in
\end{figure}

\paragraph{The hyper-parameters of ActiveLR}
To set the operations for $\alpha_{low}$ and $\alpha_{high}$, we follow \cite{tieleman2012lecture}--multiply 
by $\alpha_{low}$ and add $\alpha_{high}$. We carry out an ablation analysis of the choice of operations 
(e.g., addition or multiplication). For the results, please refer to \ref{appendix:lrlowhigh}.
To obtain the default values of $\alpha_{low}$ and $\alpha_{high}$, we carry out a preliminary experiment 
on CIFAR-10 and find that any combination of $\alpha_{low}$ and $\alpha_{high}$ that satisfy the constraint 
$\alpha_{low} + \alpha_{high} = 1$ reduce sensitivity to initial learning rates and mini-batch size, 
while increasing the overall accuracy. For all the experiments that follow, we have kept $\alpha_{low}=0.9$ and 
$\alpha_{high}=0.1$ constant. The results show the robustness of the default values of $\alpha_{low}$ and $\alpha_{high}$ 
across datasets and tasks, which suggests that ActiveLR's hyper-parameters do not need to be tuned.

\subsection{Distribution of gradient norms across layers} \label{gradNorms}
We examine the L1-norm of the gradients across layers for the optimization of the CIFAR-10 dataset on the ResNet-18 architecture, 
which has $21$ 2D-convolution layers and $1$ final fully connected layer (Figure \ref{fig:gradLayers}). The gradient norm of 
the final dense layer (layer $\# = 22$) directly correlates with the loss. In contrast, to achieve minimal loss, 
the gradient norm of the convolutional layers (layer $\# \in  [1, 21]$) increases to a maximum value (the learning phase) 
and then decreases (the stabilizing phase around the optimum).
For vanilla Adam, when the $\alpha$ is too small ($\alpha=10^{-5}$), the gradient norm for every convolutional layer 
increases monotonically with the number of iterations and the loss reaches a plateau, indicating getting stuck in a local minimum 
(\subref{fig:gradLayersA5}). In other words, when the learning rate is lower than an optimum value, the learning phase never ends. 
When the learning rate is optimal ($\alpha=10^{-3}$), Adam increases the gradient norms to an optimal level where it achieves 
the minimal loss (the end of the learning phase). Afterwards, in the stabilizing phase, the gradients are decreased at a relatively 
slow rate, which causes the loss to oscillate ($iteration>60$) around the minimum value (\subref{fig:gradLayersA3}). 
ActiveAdam, with the same low learning rate that gets Adam stuck ($\alpha=10^{-5}$), is able to achieve what Adam with 
the optimal learning rate does in the training phase (i.e., increase the gradient norms to a maximum value). 
Additionally, after the minimal loss is achieved, ActiveAdam sharply reduces the gradient norms, preventing fluctuations 
around the optimum ($iteration>60$) (\subref{fig:gradLayersC5}).

\subsection{ActiveLR and non-convex functions}
The Active implementation of a given algorithm such as Adam (ActiveAdam) provides numerous advantages over the vanilla 
implementation in troublesome non-convex optimization scenarios.

\begin{figure}[ht]
    \vskip 0.2in
    \begin{center}
    \begin{subfigure}[t]{0.3\columnwidth}
        \centerline{\includegraphics[width=\columnwidth]{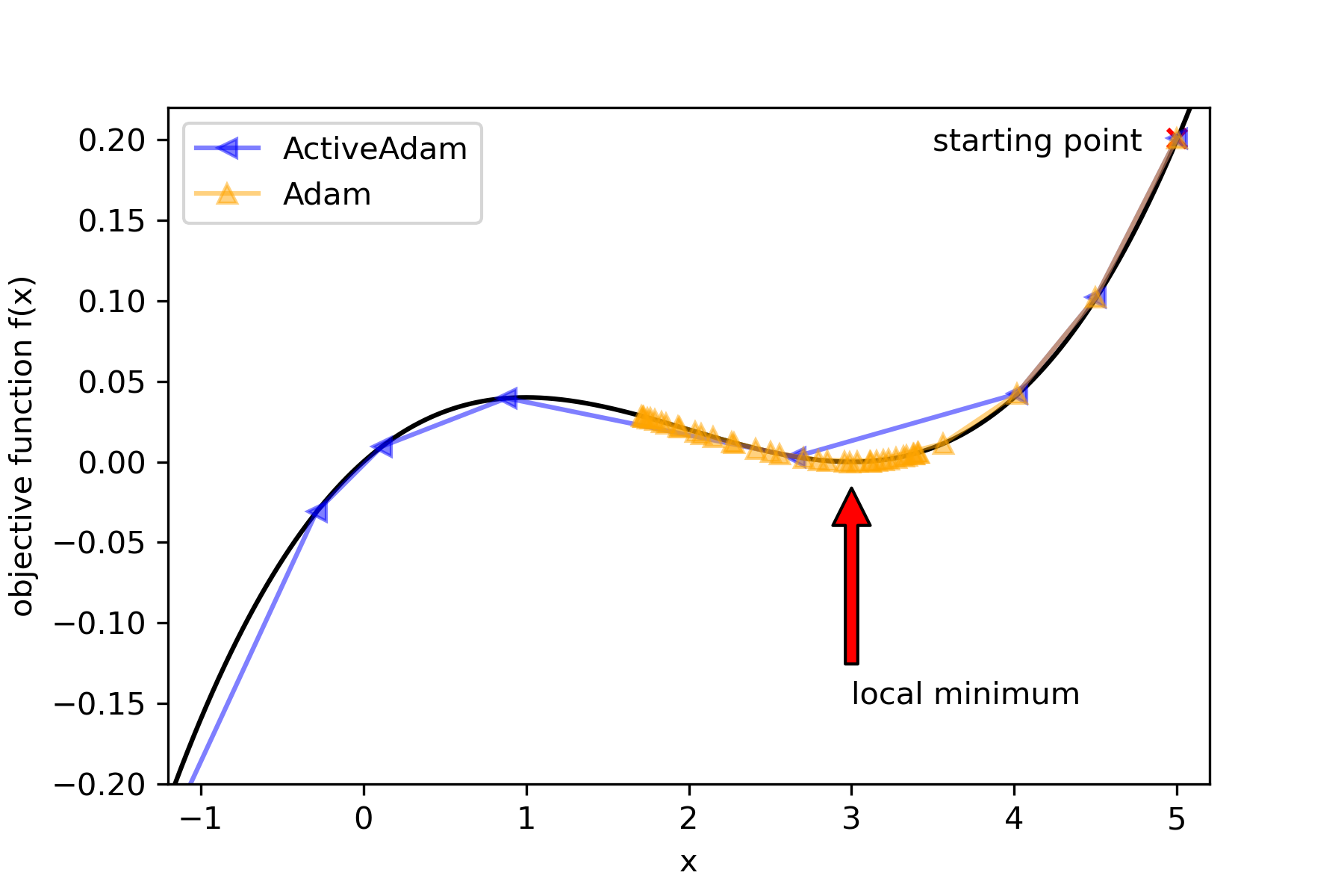}}
        \caption{$f(x)=x^3-6x^2+9x$ w.r.t. $x$}
      \label{fig:local_mina}
    \end{subfigure}
    \hfill
    \begin{subfigure}[t]{0.3\columnwidth}
        \centerline{\includegraphics[width=\columnwidth]{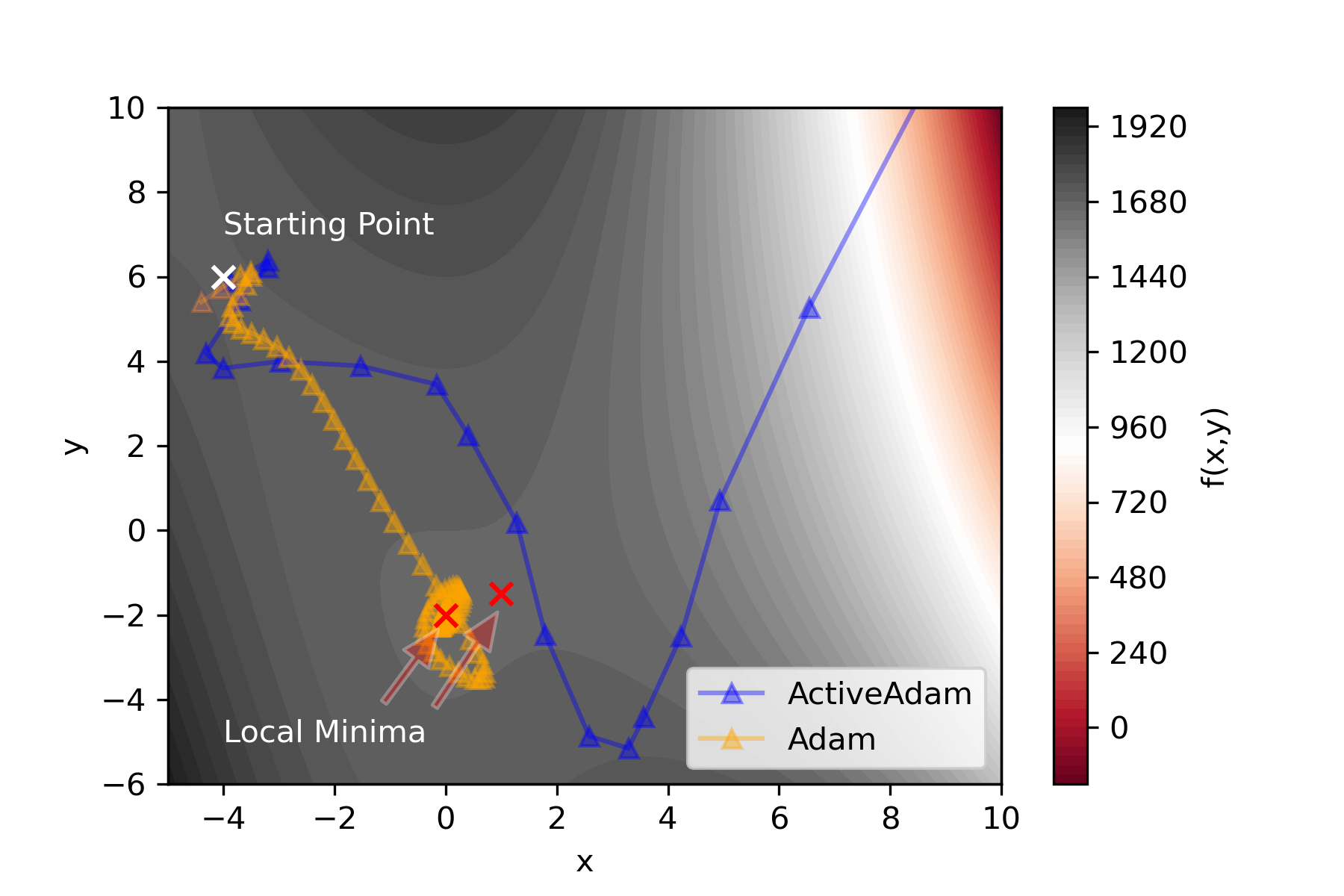}}
      \caption{$f(x,y)=-x^3-x^2y+y^2+4y+1680$ w.r.t. $x, y$}
      \label{fig:local_minb}
    \end{subfigure}
    \hfill
      \begin{subfigure}[t]{0.3\columnwidth}
        \centerline{\includegraphics[width=\columnwidth]{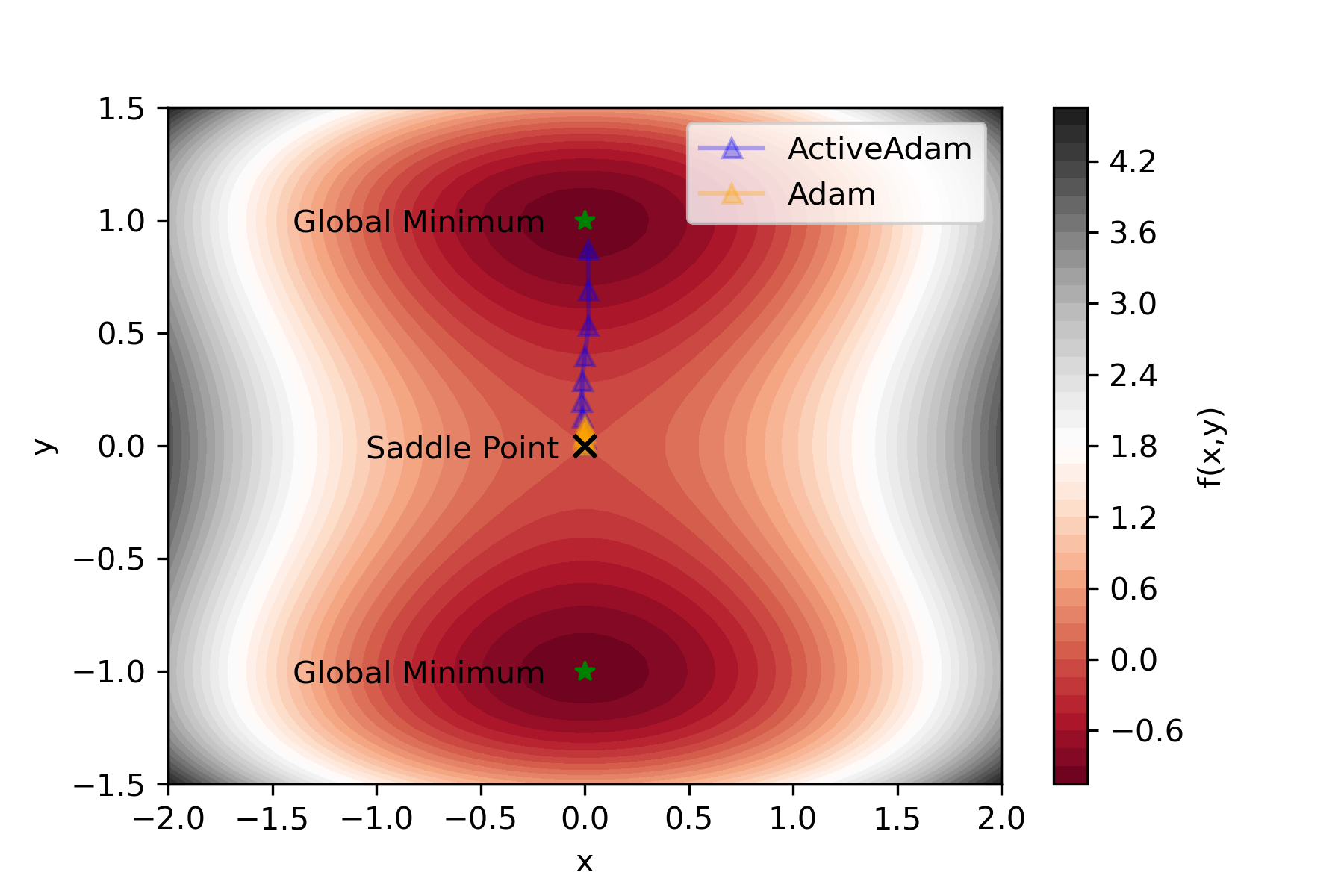}}
      \caption{$f(x, y) = y^4 - 2y^2 + x^2$ w.r.t. $x, y$}
      \label{fig:saddle}
    \end{subfigure}
  
  \caption{(\subref{fig:local_mina}) univariate unimodal, (\subref{fig:local_minb}) bivariate multimodal, and (\subref{fig:saddle}) saddle function optimization using vanilla Adam and ActiveAdam}
  \label{fig:local_min}
  \end{center}
  \vskip -0.2in
  \end{figure}

\paragraph{ActiveLR and local minima}
Local minima are a major plague in deep neural network optimization \citep{ding2019spurious, yun2018small}. 
They are one of the main reasons that optimizers do not generalize well and fail to fit the training data properly. 
This is especially the case when the learning rate is smaller than an optimum value. In Figure \ref{fig:local_mina}, 
the vanilla Adam quickly goes down the objective function at the starting point $x=5$, but as soon as it reaches the 
local minimum, $x=3$, the gradient becomes significantly small, causing vanilla Adam to oscillate around the local minimum 
no matter how long we keep training. ActiveAdam, on the other hand, quickly increases the learning rate, as soon as it 
realizes that the gradient signs are not changing. It enables ActiveAdam to pass through the local minimum and move 
towards the global minimum, $x \to -\infty$. In Figure \ref{fig:local_minb}, the vanilla optimizer faces two challenges. 
This bivariate function, $f(x,y) = -x^3-x^2y+y^2+4y+1680$, has three local minima and a global minimum of $(+\infty, +\infty)$. 
The first local minimum, $(-4,6)$, the starting point, has a higher function value than the other two, $(0,-2)$ 
and $(1, -\frac{3}{2})$, which have the same lower value. The first challenge is starting near a local minimum, 
where the gradients are too small. Adam struggles to escape the starting point because of the small gradients. 
After a large number of iterations, Adam eventually escapes the starting local optimum but faces the second challenge. 
It gets trapped in another local minimum, $(0,-2)$, and is unable to escape. ActiveAdam, on the other hand, 
quickly escapes the initial local minimum since it increases its learning rate when the gradients do not change. 
With the higher learning rate it has accumulated, it is able to quickly escape the other two local minima and converge 
to the global minimum, $(+\infty, +\infty)$.

\paragraph{ActiveLR and saddle points}
Saddle points are another major pitfall in training neural networks \citep{kawaguchi2016deep}. 
For deep neural networks, large sets of strict and non-strict saddle points have been shown to exist \citep{achour2021global}. 
Being close to a saddle point equates very small gradient values in all directions. 
As a result, non-Active optimizers will require significantly higher training iterations to escape saddle points.
In Figure \ref{fig:saddle}, the saddle point lies at $(0,0)$. After $12$ iterations, Adam is relatively
where it was when the training started. ActiveAdam, however, quickly increases its learning rate, 
following iterations of no sign change, and reaches the global minimum at $(0,1)$ in $12$ iterations.

\begin{figure}[ht]
    \vskip 0.2in
    \begin{center}
        \centerline{\includegraphics[width=\columnwidth]{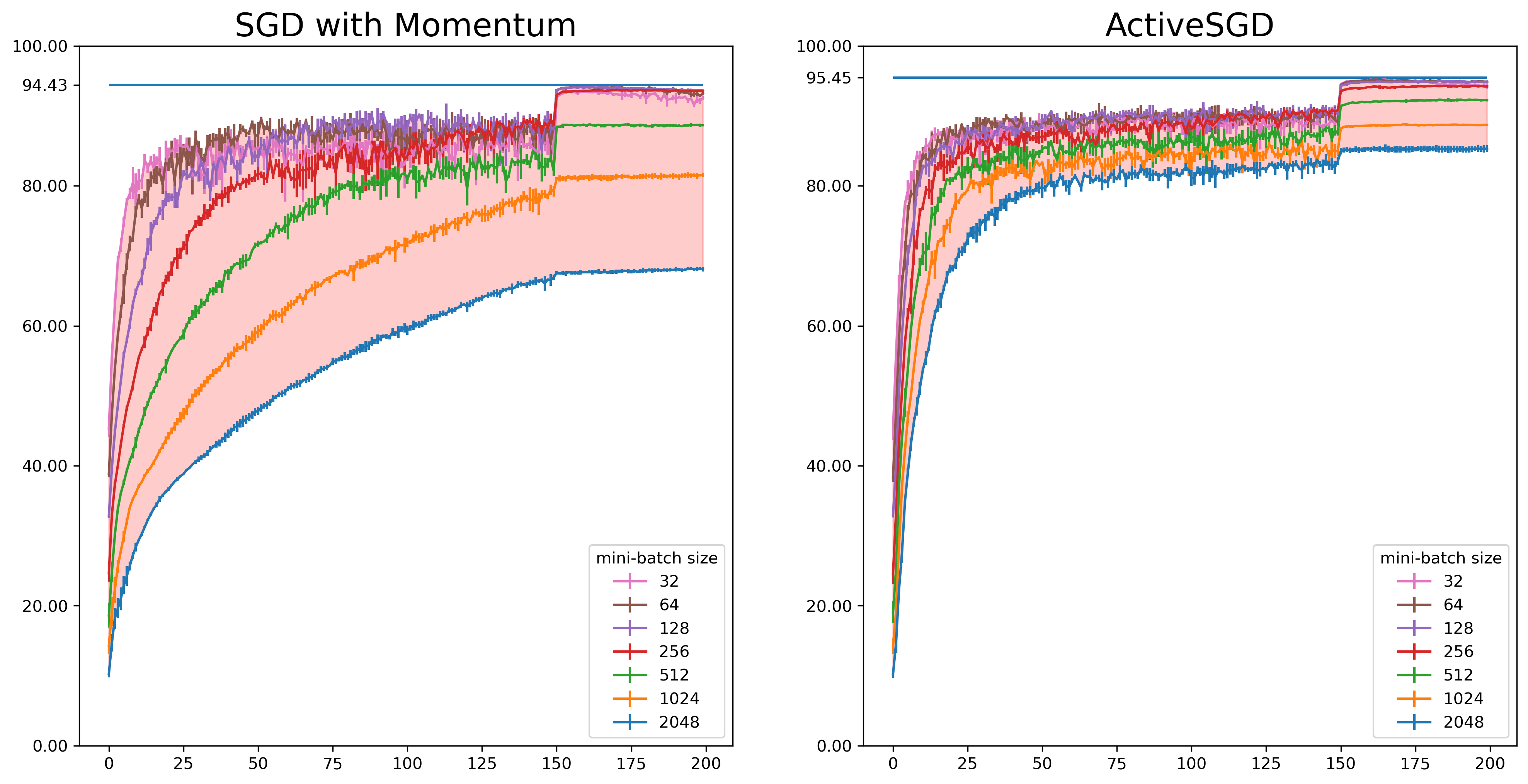}}
    \caption{Reduced sensitivity to mini-batch size on CIFAR-10 test accuracy $([\mu \pm \sigma])$}
    \label{fig:bs}
    \end{center}
    \vskip -0.2in
  \end{figure}

  \section{Experiments}\label{experiments}
  We perform our experiments on ImageNet \citep{deng2009imagenet} with ResNet-18 and CIFAR-10 
  \citep{krizhevsky2009learning} with ResNet-34 for image classification, WikiText-2 and WikiText-103 
  \citep{merity2016pointer} with GPT-2 \cite{radford2019language} for language modelling, and PASCAL VOC 
  \cite{Everingham10} dataset on Faster-RCNN+FPN for object detection. Based on the literature, 
  non-adaptive optimizers (e.g. SGD) work with higher initial learning rates compared to adaptive optimizers. 
  Therefore, for all the experiments, we use $50$ times the hyperparameter space for initial learning rates of SGD 
  with momentum and ActiveSGD compared to the adaptive optimizers. For each vanilla optimizer and its Active variant, 
  we test the performance across a range of initial learning rates and, for other hyperparameters, we use the values 
  suggested on the optimizer's official GitHub page for each task.
  We use a heterogeneous cluster with each node comprised of 4 x NVIDIA P100 Pascal (12G or 16G HBM2 memory) 
  or 4 x NVIDIA V100 Volta (32G HBM2 memory). To simulate common practitioners' limited access to GPUs, 
  we train the WikiText-2, WikiText-103, PASCAL VOC, and CIFAR-10 experiments on 1 GPU and the ImageNet experiments on 4 GPUs.
  
  \begin{figure}[ht]
    \vskip 0.2in
    \begin{center}
        \centerline{\includegraphics[width=\columnwidth]{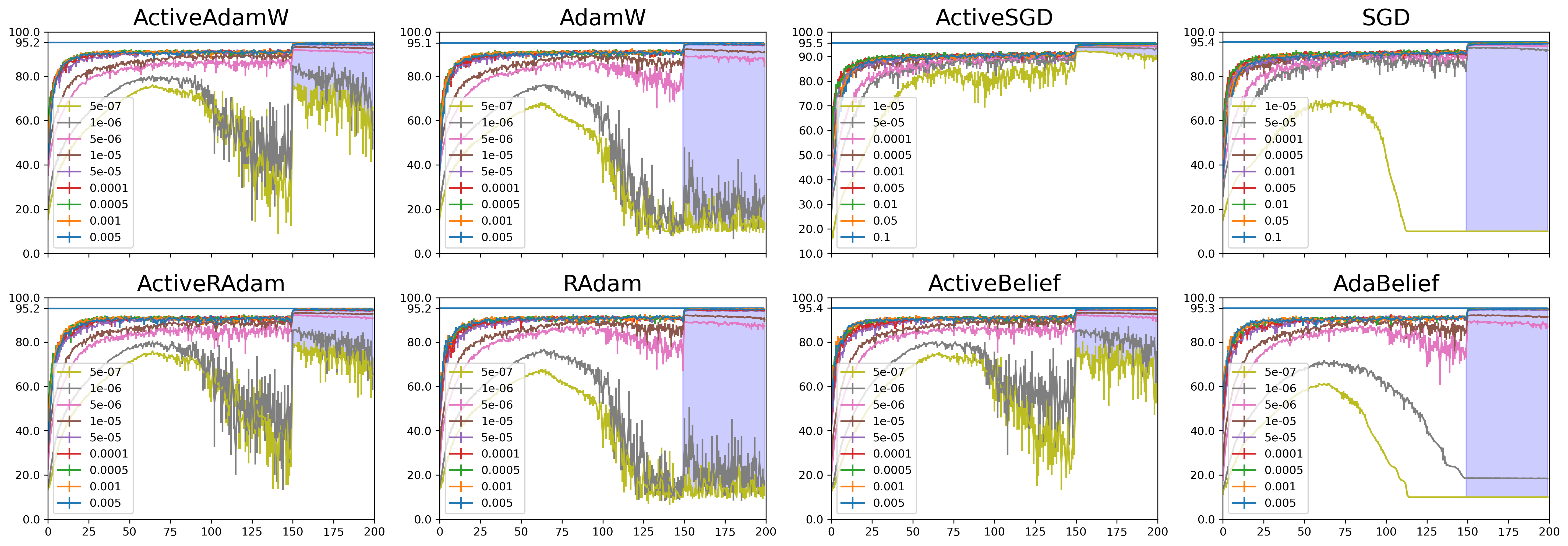}}
    \caption{Test accuracy $([\mu \pm \sigma])$ of SGD with momentum, AdaBelief, RAdam, and AdamW compared with their Active variant on CIFAR-10. The red area in-between lines indicates the sensitivity of the optimizer w.r.t. the value of the initial learning rate. The horizontal lines indicate the highest mean accuracy of each optimizer across learning rates.}
    \label{fig:cifar10}
    \end{center}
    \vskip -0.2in
  \end{figure}
  
\subsection{Sensitivity to mini-batch size}
ActiveLR alleviates the problem of large mini-batch sizes by its automatic learning rate adaptation. 
In Figure \ref{fig:bs}, we can see that for the CIFAR-10 dataset, increasing the mini-batch size from the optimal value of $128$ 
to larger values significantly decreases the test-set accuracy of vanilla SGD with momentum. Larger mini-batch sizes 
also destabilize training as can be seen by higher fluctuations in accuracy for vanilla SGD with momentum. ActiveSGD, 
on the other hand, achieves a relatively high accuracy regardless of the mini-batch size and also remains stable. 
($\alpha={10}^{-4}$, ResNet-18)
\subsection{Sensitivity to learning rate}
\paragraph{CIFAR-10} \label{cifar_exp}
We train SGD with momentum, AdaBelief, RAdam, and AdamW along with their Active variant 
(i.e., ActiveSGD, ActiveBelief, ActiveRAdam, ActiveAdamW) with 9 different initial learning rates 
$5 \times 10^n$, $n \in [-7,-3]$, each ran $3$ times. We test different values for weight decay in the range of 
$[10^3,10^{-8}]$ and found that the optimal weight decay, $wd^*$ depends on the learning rate, $\alpha$. 
For CIFAR-10, ${wd}^* \times  \alpha = 10^{-4}$.
As seen in Figure \ref{fig:cifar10}, the Active variant of each optimizer achieves a higher average accuracy 
compared with the vanilla variant. More importantly, the area between the accuracy achieved by the worst-performing 
learning rate and the best-performing learning rate is significantly smaller for the Active variants compared to the 
vanilla variants. This indicates reduced sensitivity to the selection of the initial learning rate that ActiveLR achieves.
  
\begin{figure}[ht]
    \begin{center}
    \begin{subfigure}[t]{.49\columnwidth}
        \centerline{\includegraphics[width=\columnwidth]{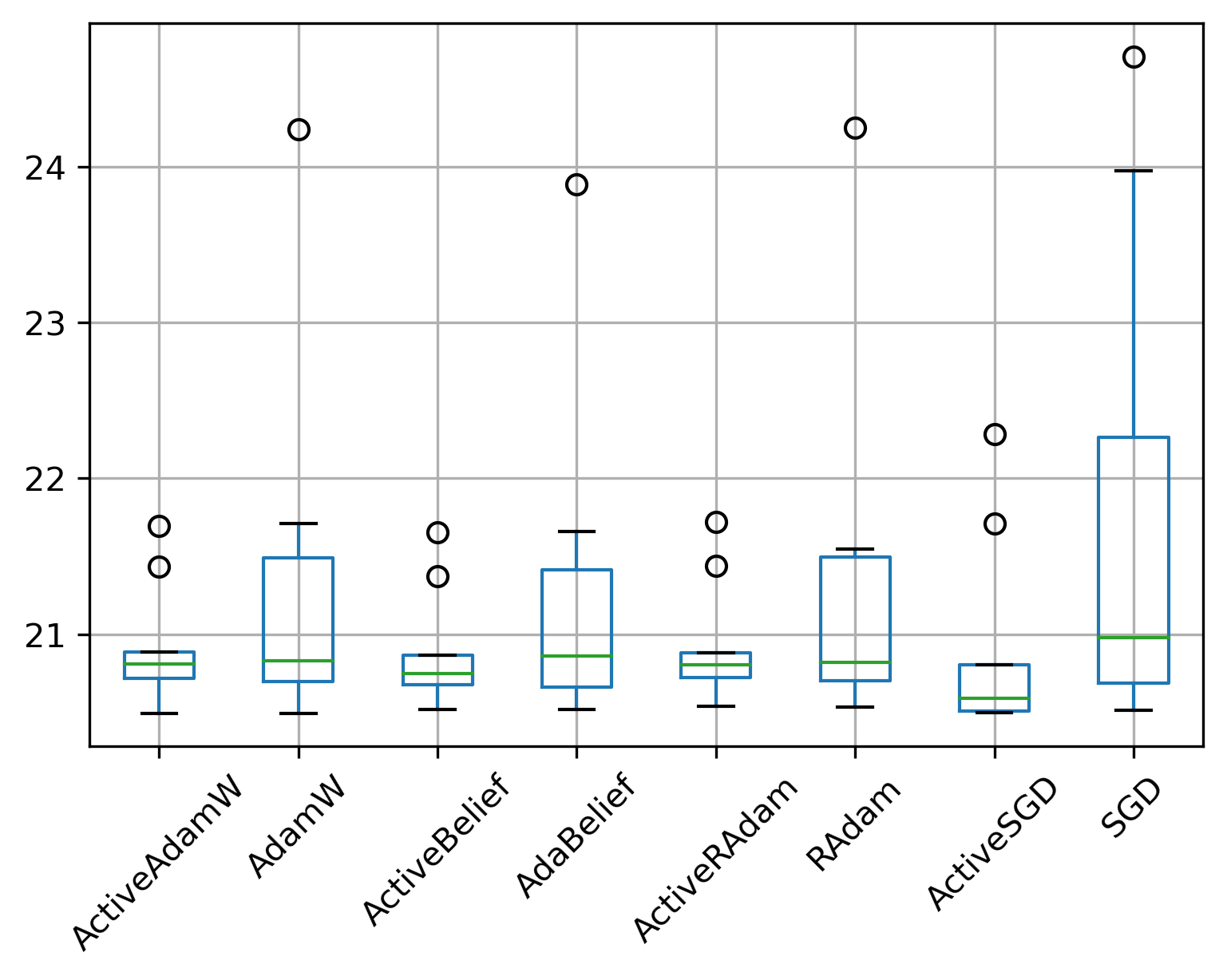}}
          \caption{Test-set PPL (lower is better)}
    \end{subfigure}
    \hfill
      \begin{subfigure}[t]{.49\columnwidth}
        \centerline{\includegraphics[width=\columnwidth]{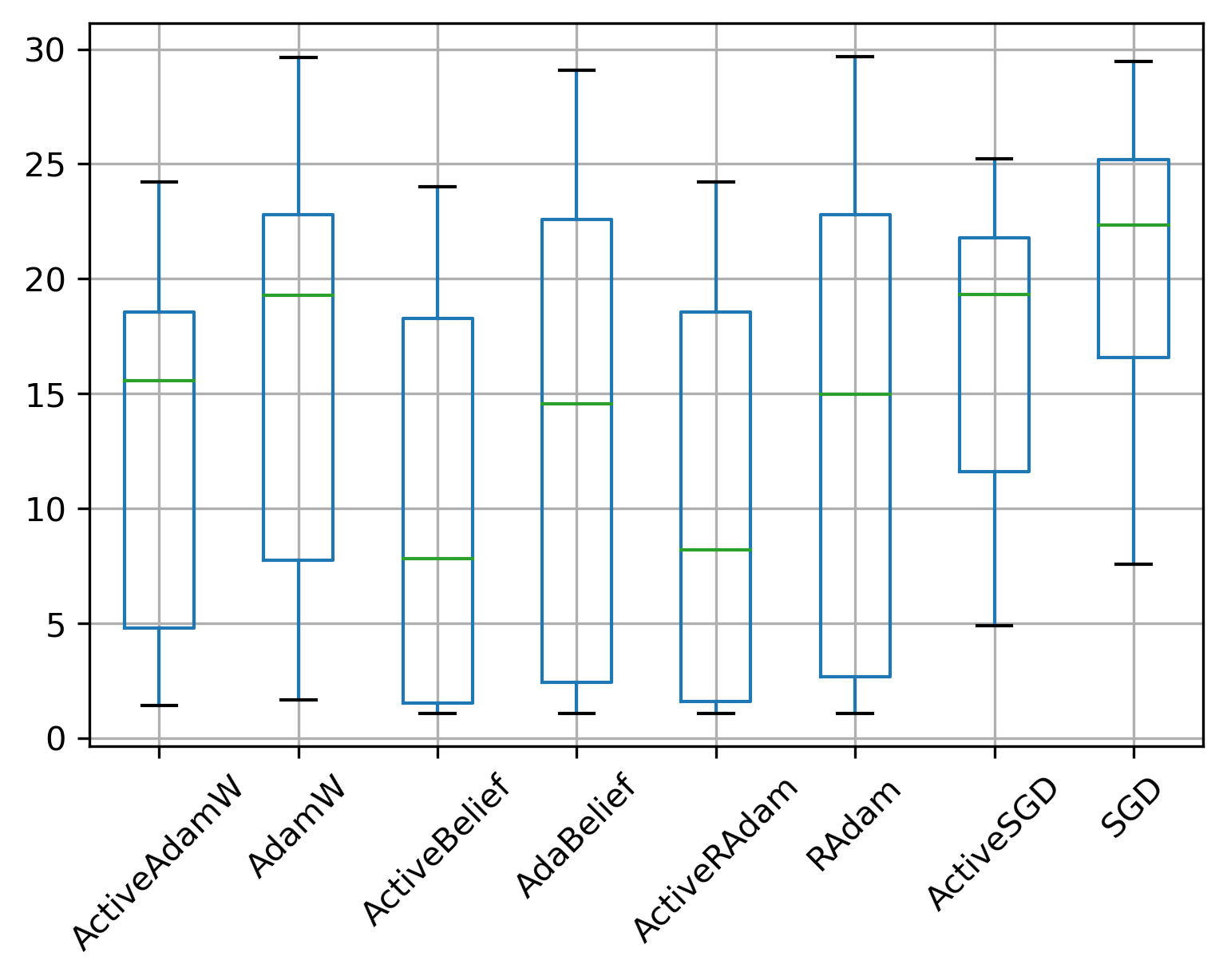}}
        \caption{Train-set PPL (lower is better).}
    \end{subfigure}
    \caption{WikiText-2 on GPT-2 with various initial learning rates}
    \label{fig:wiki}
    \end{center}
    \vskip -0.2in
  \end{figure}
  
  \paragraph{WikiText-2}\label{wikitext_exp}
  We fine-tune GPT-2 with 124,439,808 parameters on WikiText-2 (vocabulary size $50257$, maximum sequence length $1024$, 
  dimensionality of the embeddings and hidden states $768$, number of hidden layers in the Transformer encoder $12$, 
  number of attention heads for each attention layer $12$, activation function $gelu$, dropout probability $0.1$). 
  Initial learning rates in the range of $[10^{-8},10^{-4}]$ for adaptive optimizers and $[5 \times 10^{-7}, 5 \times 10^{-3}]$ 
  for non-adaptive optimizers are tested. For each initial learning rate, we calculate the test-set PPL for the epoch 
  that gives the highest validation-set PPL. In Figure \ref{fig:wiki}, we can see that ActiveLR gives better test-set 
  PPL compared to vanilla optimizers on average, while producing lower train-set errors. Furthermore, vanilla optimizers 
  have a significant degree of variation in their test-set and train-set PPLs, indicating their sensitivity to the initial 
  learning rate, while ActiveLR shows relative insensitivity to the initial learning rate.
  A surprising finding is the generalizability of ActiveSGD. Although in the literature the performance of SGD with momentum 
  for transformers models has been shown to be inferior to adaptive methods--and our results confirm it--the ActiveLR 
  variant of SGD with momentum shows better generalizability than the adaptive methods. In fact, ActiveSGD gives the best 
  test PPL among all optimizers tested.
  
  
  \begin{figure}[ht]
    \vskip 0.2in
    \begin{center}
    \begin{subfigure}[t]{.3\columnwidth}
        
        \centerline{\includegraphics[width=\columnwidth]{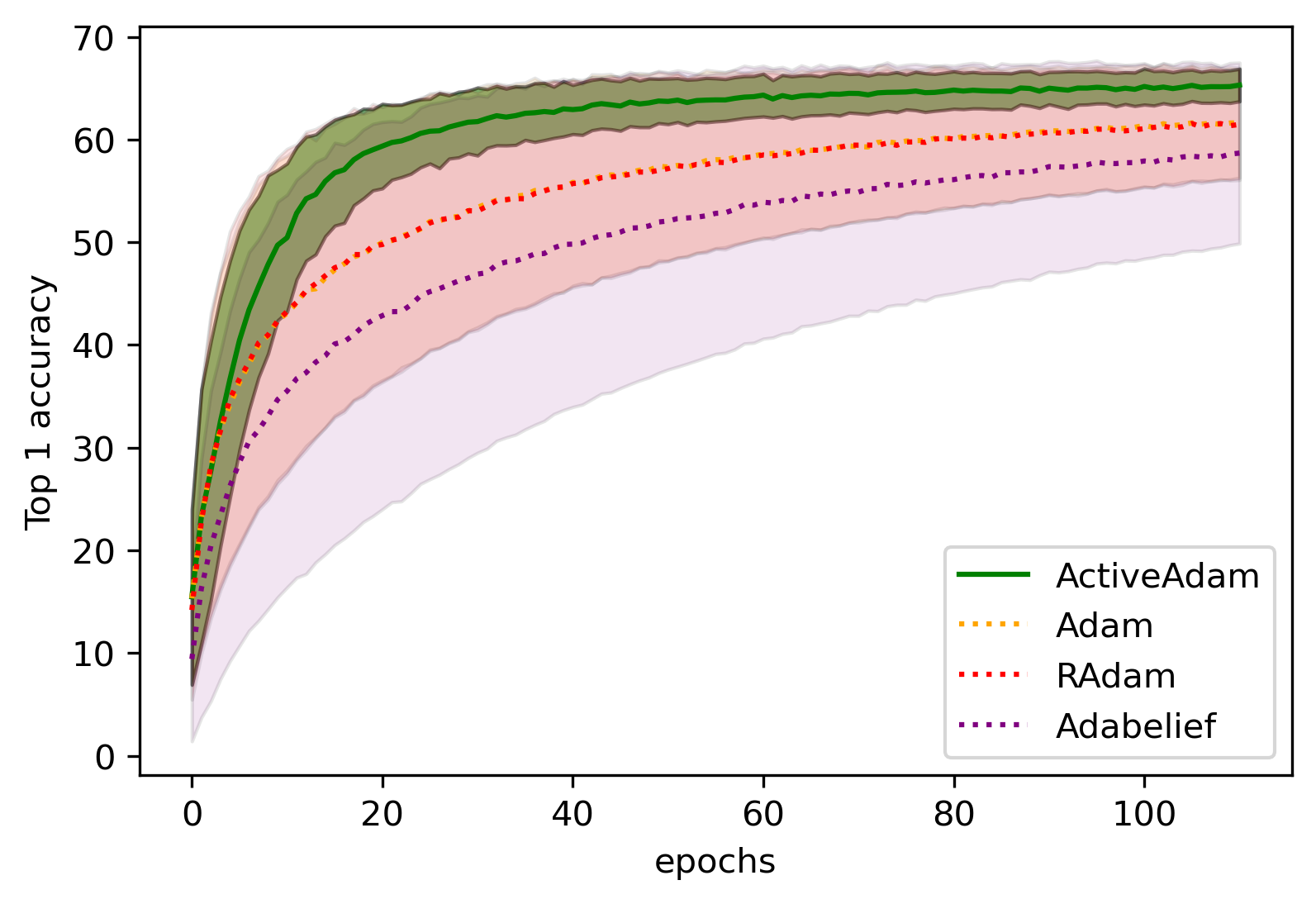}}
        \caption{ImageNet top 1 accuracy \\$([\mu \pm \sigma])$}
    \end{subfigure}
    \begin{subfigure}[t]{.3\columnwidth}
        
        \centerline{\includegraphics[width=\columnwidth]{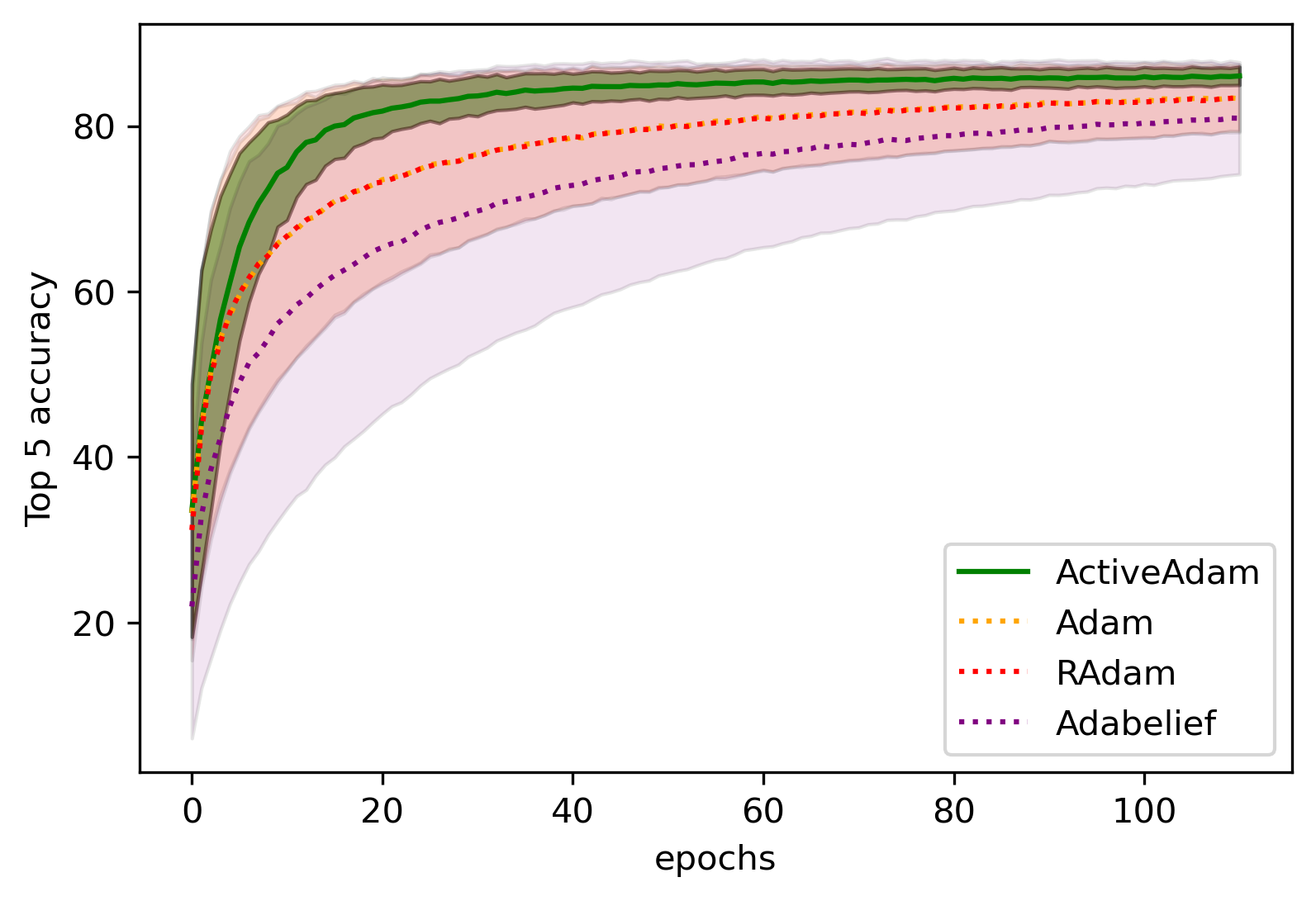}}
        \caption{ImageNet top 5 accuracy \\$([\mu \pm \sigma])$}
    \end{subfigure}
    \begin{subfigure}[t]{.3\columnwidth}
        \centerline{\includegraphics[width=\columnwidth]{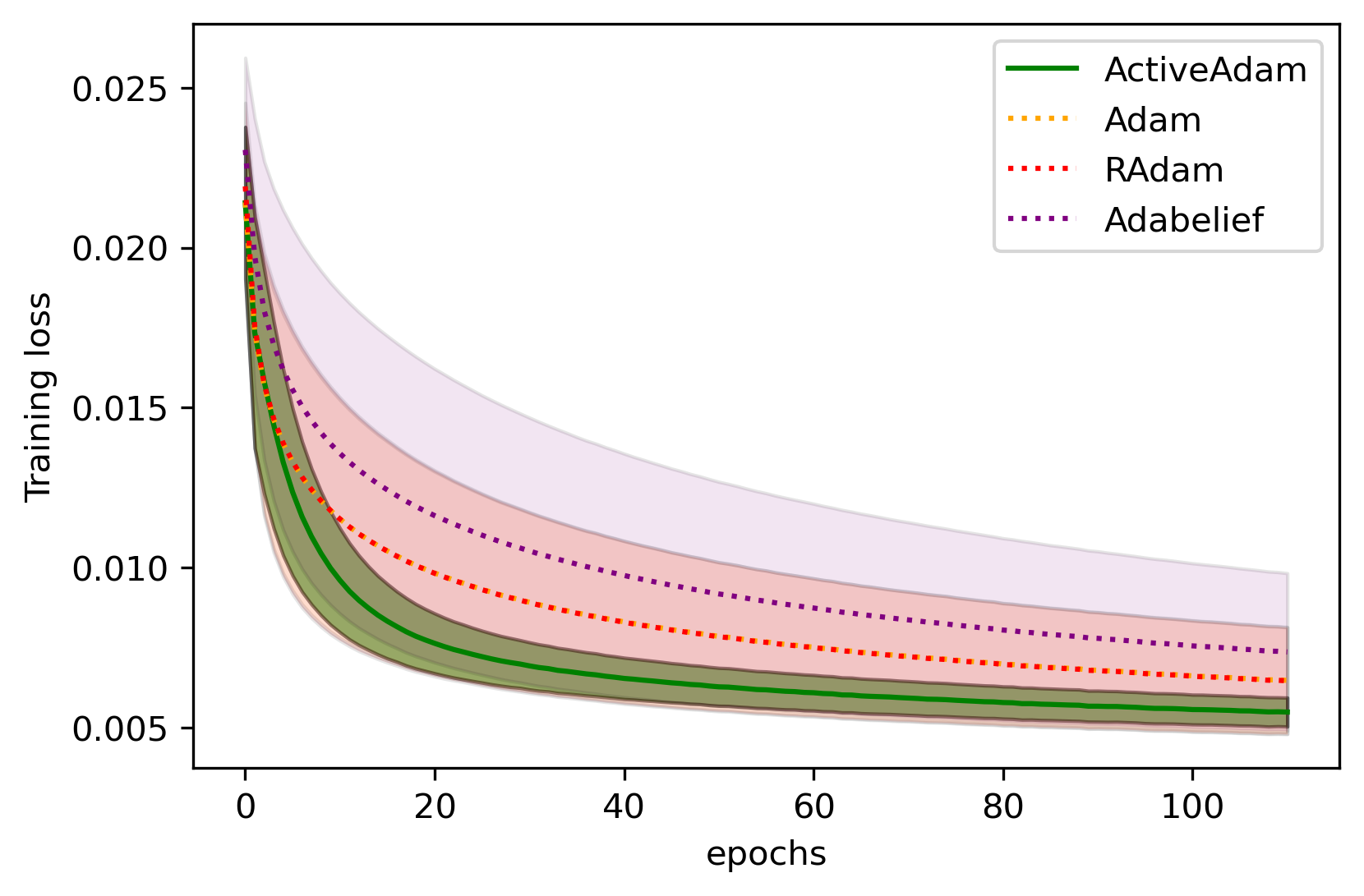}}
        \caption{ImageNet training loss \\$([\mu \pm \sigma])$}
    \end{subfigure}
    \caption{The mean/std test accuracy $([\mu \pm \sigma])$ of ActiveAdam, AdaBelief, RAdam, and Adam on ImageNet with three initial learning rates ($[10^{-5},10^{-4},10^{-3}]$). The shaded area around each line is the standard deviation of the means for each optimizer across different values of the initial learning rate. (Adam and RAdam plots are \emph{visually identical})}
    \label{fig:imagenet}
    \end{center}
    \vskip -0.2in
  \end{figure}
  
  \paragraph{ImageNet}\label{imagenet_exp}
  For ImageNet, we train ActiveAdam, AdaBelief, RAdam, and Adam for $110$ epochs, after which the accuracies do not improve.
  Figure \ref{fig:imagenet} shows the average and standard deviation of each metric across initial learning rates of ${10}^n, n \in \{-3,-4,-5\}$. ActiveAdam achieves the highest mean top-1 and top-5 accuracy and the lowest training error (the green line) compared to the other vanilla optimizers (the plots for Adam and RAdam are \emph{visually identical}). It also has the lowest amount of variance among the optimizers, showing reduced sensitivity to the initial learning rate.
  \paragraph{WikiText-103 and PASCAL VOC} Please refer to Appendix \ref{extraexp}.
  \section{Social impact}
  ActiveLR makes contributions to several social causes. First, given the growth of the size of datasets and deep learning models, it has become increasingly more difficult for average AI practitioners and researchers to train state-of-the-art models. By reducing sensitivity to hyperparameter initialization, ActiveLR helps in democratizing AI. Second, carbon emissions from training deep neural networks have reached alarming levels. Consider training a new dataset with the size of the ImageNet on a relatively small model, such as ResNet-18. Let us assume the goal is to achieve top-1 accuracy of 55\% or higher. Since we a priori do not know what the optimal initial learning rate is, we need to test various initial learning rates. Testing 9 initial learning rates starting from $10^{-4}$ up to ${10^0}$, we find that, with SGD with momentum, 7 out of 9 initial learning rates achieve our desired accuracy in $100$ epochs, while, with ActiveSGD, 8 out of 9 values reach our desired accuracy (Figure \ref{fig:imagenetsgd}).
  Considering the worst-case, where the suboptimal learning rates are selected first, SGD with momentum, compared with ActiveSGD, requires 13,816,000 TFLOPS extra computation (2200 seconds per epoch, 61.11 hours total). With an average carbon efficiency of 0.432 kgCO$_2$eq/kWh, using 4 x Tesla V100-SXM2-32GB (TDP of 300W) GPUs (same as our experiments), the excessive $CO_2$ emissions are estimated to be 7.92 kgCO$_2$eq. This is equivalent to 3.96 Kgs of extra coal burned. On the other hand, ActiveSGD not only prevents such extra carbon emissions, but also gives higher average accuracy and fit (estimations were conducted using the \href{https://mlco2.github.io/impact#compute}{MachineLearning Impact calculator} presented in \citet{lacoste2019quantifying}.) As a result, we believe that ActiveLR is a step towards eliminating the need for hyperparameter search, which helps democratize AI and reduce carbon emissions.
  
  \section{Limitations}\label{limitations}
  While we try to include as many experiments on various tasks to show the robustness of the results of ActiveLR, due to time and computational limits, we could not test all benchmark datasets. We are working on releasing test results for COCO dataset \cite{lin2014microsoft} for object segmentation, and IWSLT14 DE-EN for neural machine translation on transformers, among others, for ActiveLR.
  With Active LR, we demonstrated the capability to reduce sensitivity to initial learning rate and mini-batch size. We encourage future research to identify and tackle optimization sensitivity to \emph{other} hyper-parameters, such as weight decay, in an attempt to streamline neural network optimization, reduce carbon emissions, and training time.
  \section{Conclusion}
  In this paper, we show that for all the learning rates tested, the Active variant of SGD with momentum, AdamW, AdaBelief, and RAdam achieves the best fit and highest accuracy. Our tests show robust results for different datasets and model architectures. It did so in a lower number of epochs, which translates to faster training. Moreover, we show that ActiveLR remains stable and does not suffer from loss in generalizability when trained with large mini-batch sizes. As a result, multi-GPU training of large datasets can be sped up with ActiveLR using large mini-batch sizes. The orthogonality of ActiveLR to other optimizers allows it to be implemented on top of them.
  We also show relative insensitivity of ActiveLR to the values of initial learning rate and mini-batch size. We encourage practitioners to use ActiveLR for their model training to achieve better training/test-set performance in a shorter amount of time, remove the need for tuning the initial learning rate and mini-batch size, and reduce their carbon footprint.

\section*{Acknowledgements}
The authors would like to thank Compute Canada for its provision of computational resources that made the experimentations possible.

\bibliography{refs}
\bibliographystyle{icml2022}

\newpage
\appendix
\section{Appendix}
\subsection{Additional plots}
\begin{figure}[ht]
    \vskip 0.2in
    \begin{center}
        \begin{subfigure}[t]{0.3\columnwidth}
  	
            \centerline{\includegraphics[width=\columnwidth]{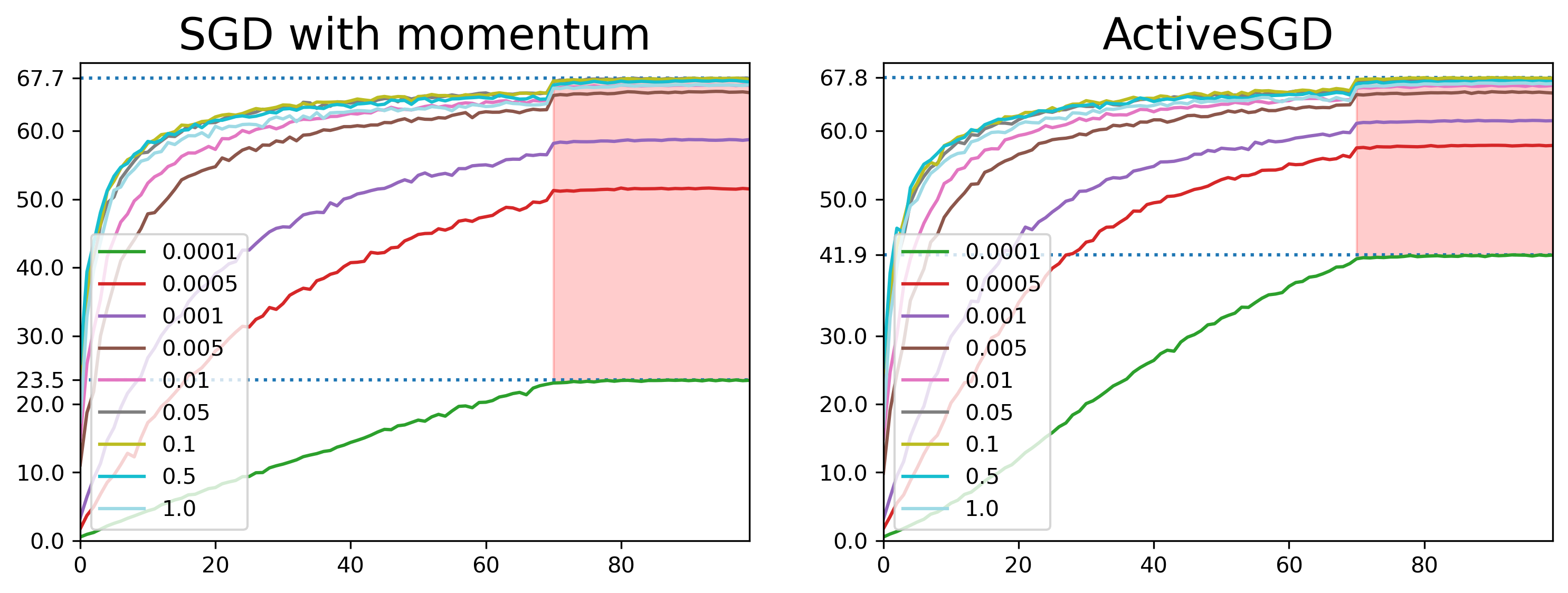}}
  	\caption{ImageNet top-1 accuracy}
  \end{subfigure}
  \begin{subfigure}[t]{0.3\columnwidth}
  	
    \centerline{\includegraphics[width=\columnwidth]{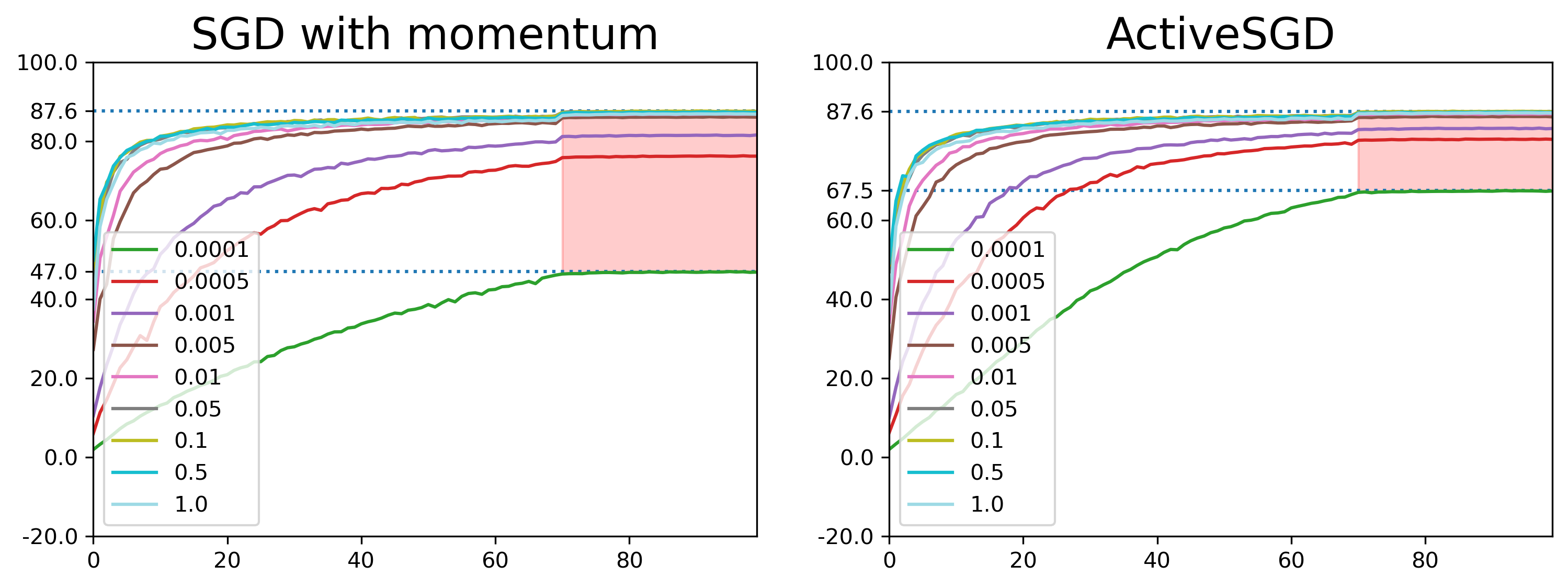}}
  	\caption{ImageNet top-5 accuracy}
  \end{subfigure}
  \begin{subfigure}[t]{0.3\columnwidth}
  	
    \centerline{\includegraphics[width=\columnwidth]{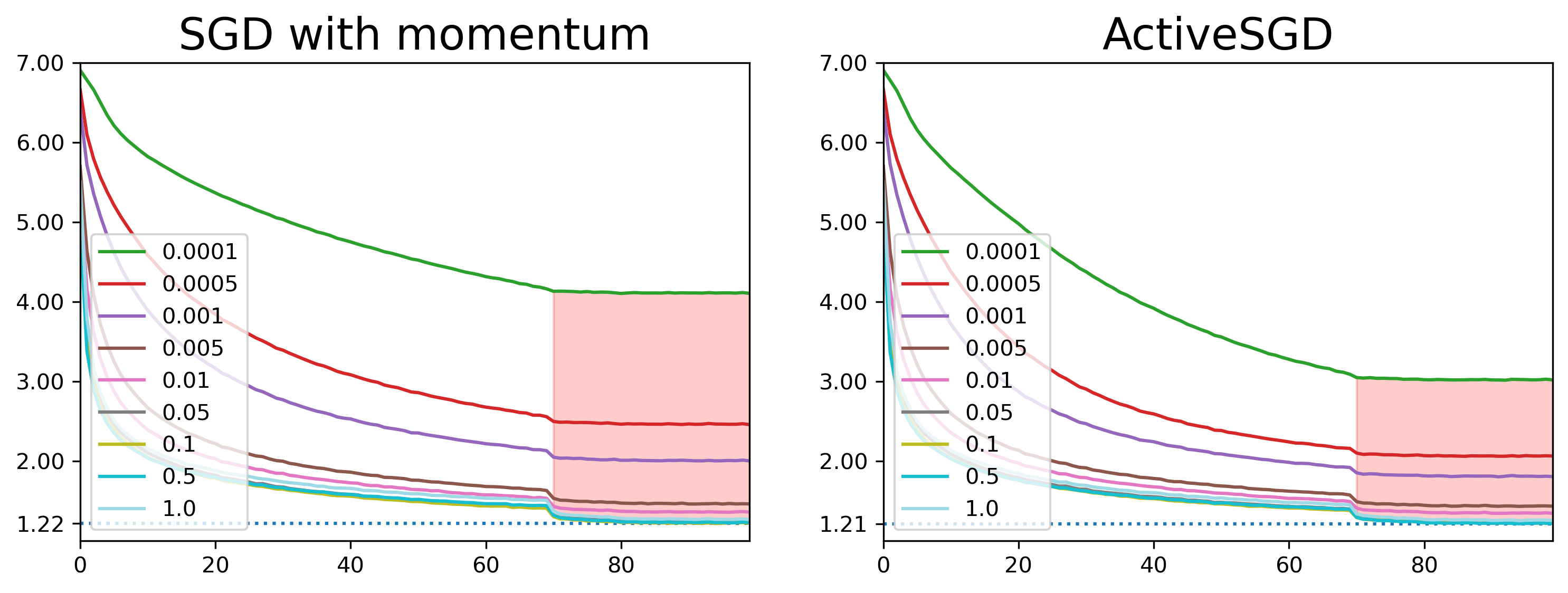}}
  	\caption{ImageNet training loss}
  \end{subfigure}
  \caption{The top-1, top-5, and training loss of SGD with momentum and ActiveSGD on ImageNet with 9 initial learning rates $[10^{-4},10^{0}]$. The shaded area shows the performance gap between the best initial learning rate and the worst one (sensitivity to the value of initial learning rate.)}
  \label{fig:imagenetsgd}
\end{center}
\vskip -0.2in
\end{figure}

SGD with momentum vs. ActiveSGD for Image-Net on ResNet-18 (Figure \ref{fig:imagenetsgd}).

\begin{figure}[ht]
    \vskip 0.2in
    \begin{center}
    \centerline{\includegraphics[width=\columnwidth]{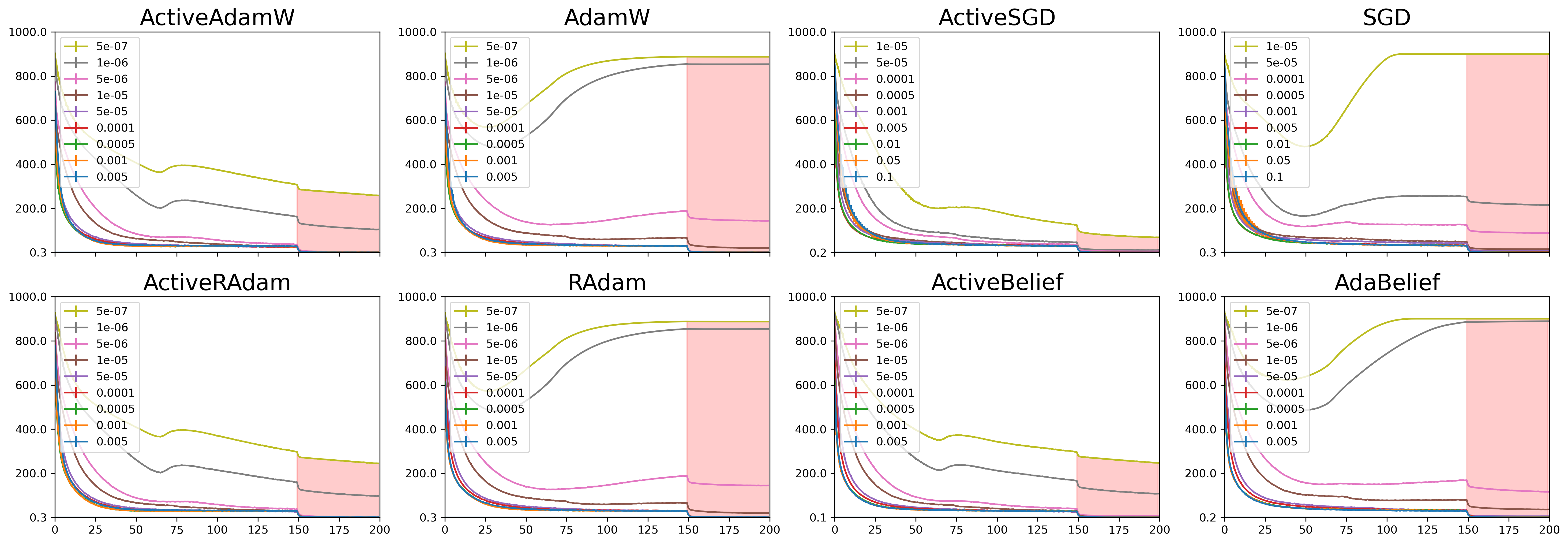}}
  \caption{Training loss $([\mu \pm \sigma])$ of SGD with momentum, AdaBelief, RAdam, and AdamW compared with their Active variant on CIFAR-10. The red area in-between lines indicates the sensitivity of the optimizer w.r.t. the value of the initial learning rate. The horizontal lines indicate the lowest loss of each optimizer across learning rates.}
  \label{fig:cifar10train}   
  \end{center}
  \vskip -0.2in
\end{figure}

CIFAR-10 training loss on ResNet-34 (Figure \ref{fig:cifar10train}).

\subsection{Extra experiments} \label{extraexp}
\paragraph{WikiText-103 on mini-GPT2}
We train WikiText-103 on a small variant of GPT-2--mini-GPT-2 \ref{minigpt2specs} from scratch.

\begin{figure}[ht]
    \vskip 0.2in
    \begin{center}
  \begin{subfigure}[t]{.49\columnwidth}

    \centerline{\includegraphics[width=\columnwidth]{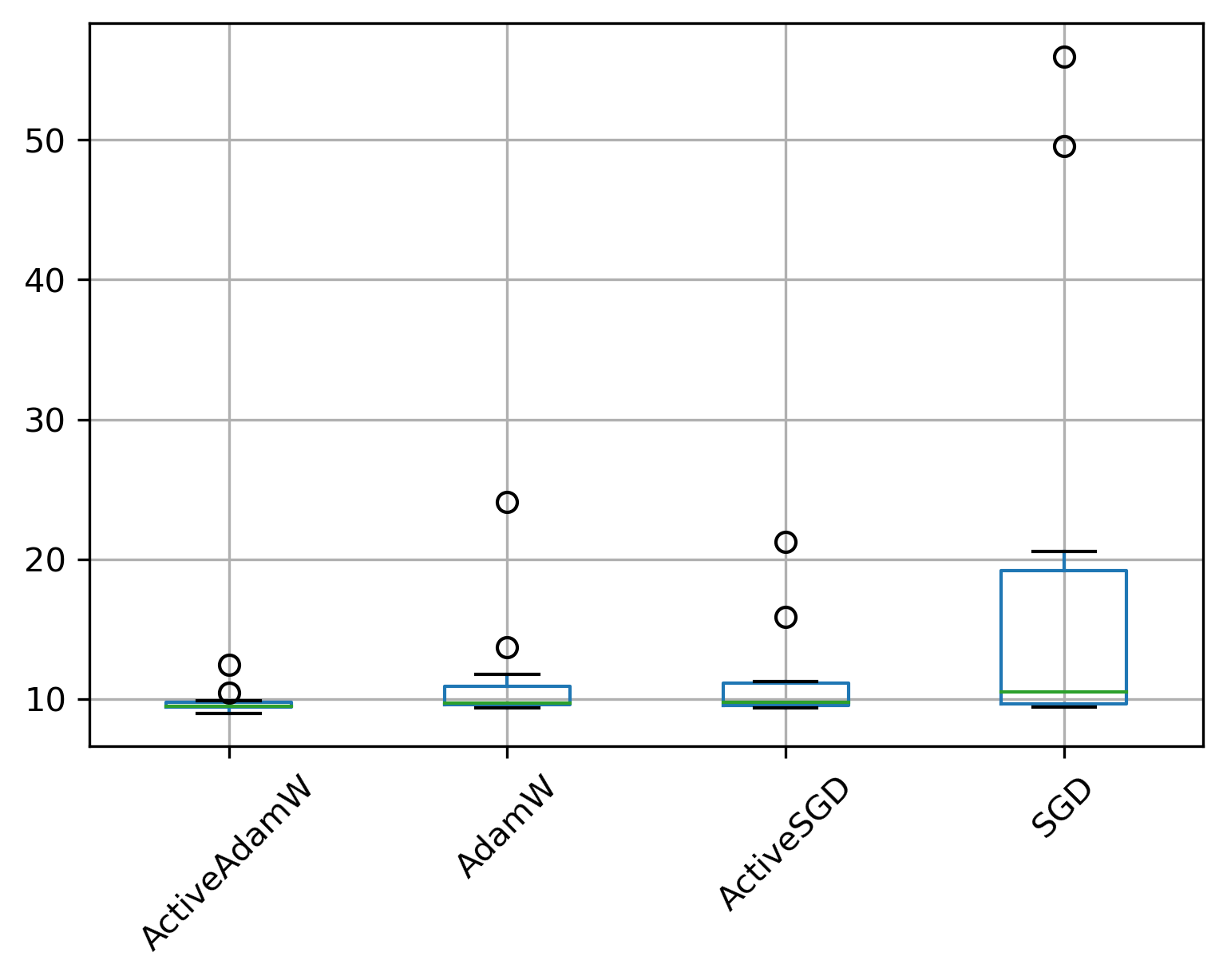}}
        \caption{Test-set PPL (lower is better)}
  \end{subfigure}%
  \hfill
  \begin{subfigure}[t]{.49\columnwidth}
  	
    \centerline{\includegraphics[width=\columnwidth]{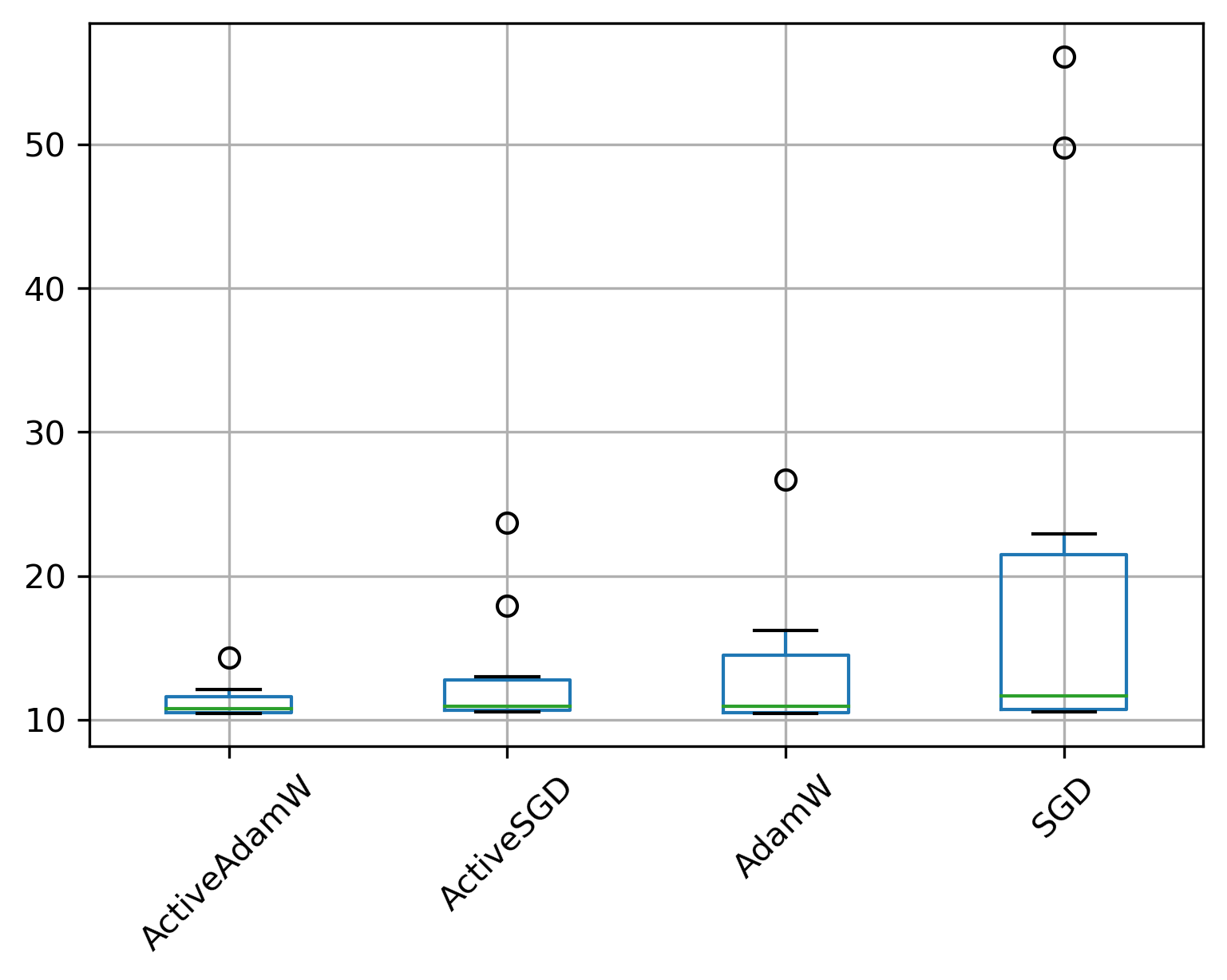}}
        \caption{Train-set PPL (lower is better)}
  \end{subfigure}
  \caption{WikiText-103 on mini-GPT-2 with various initial learning rates}
  \label{fig:wiki103}
\end{center}
\vskip -0.2in
\end{figure}

Figure \ref{fig:wiki103} shows the test and training PPLs for SGD with momentum, AdamW, ActiveSGD and ActiveAdamW. They show that compared with the vanilla variants, the ActiveLR variants achieve better training fit and test-set PPL. Moreover, lower variance of ActiveSGD and ActiveAdamW PPLs show the reduced sensitivity of the ActiveLR variants to the value of the initial learning rate compared with the vanilla variants.

\begin{table}[t]
    \caption{mini-GPT-2 specifications}
    \label{minigpt2specs}
    \vskip 0.15in
    \begin{center}
    \begin{small}
    \begin{sc}
    \begin{tabular}{lcccr}
    \toprule

      Hyperparameter     & Value \\
      \midrule
      Vocabulary Size & $10,000$     \\
      Block Size     & $200$    \\
      Number of Layers     & 2            \\
      Number of Attention Heads & 2 \\
  
      \bottomrule
    \end{tabular}
\end{sc}
\end{small}
\end{center}
\vskip -0.1in
\end{table}
  
The training hyperparameters are in Table \ref{wiki103hyper}.
  \begin{table}[t]
    \caption{WikiText-103 on mini-GPT-2 hyperparameters}
    \label{wiki103hyper}
    \vskip 0.15in
    \begin{center}
    \begin{tiny}
    \begin{sc}
    \begin{tabular}{lcccr}
    \toprule
      Hyperparameter     & Value \\
      \midrule
      Mini-batch Size & $128$     \\
      Maximum Number of Epochs     & $200$    \\
      Weight Decay     & $0.0$            \\
      Non-Adaptive Optimizers' Learning Rates & $[5 \times 10^{-6}, 10^{-1}]$\\
      Adaptive Optimizers' Learning Rates & $[10^{-7}, 10^{-2}]$\\
      
      \bottomrule
\end{tabular}
\end{sc}
\end{tiny}
\end{center}
\vskip -0.1in
\end{table}
  
\begin{figure}[ht]
    \vskip 0.2in
    \begin{center}
    \begin{subfigure}[b]{.49\columnwidth}
        
        \centerline{\includegraphics[width=\columnwidth]{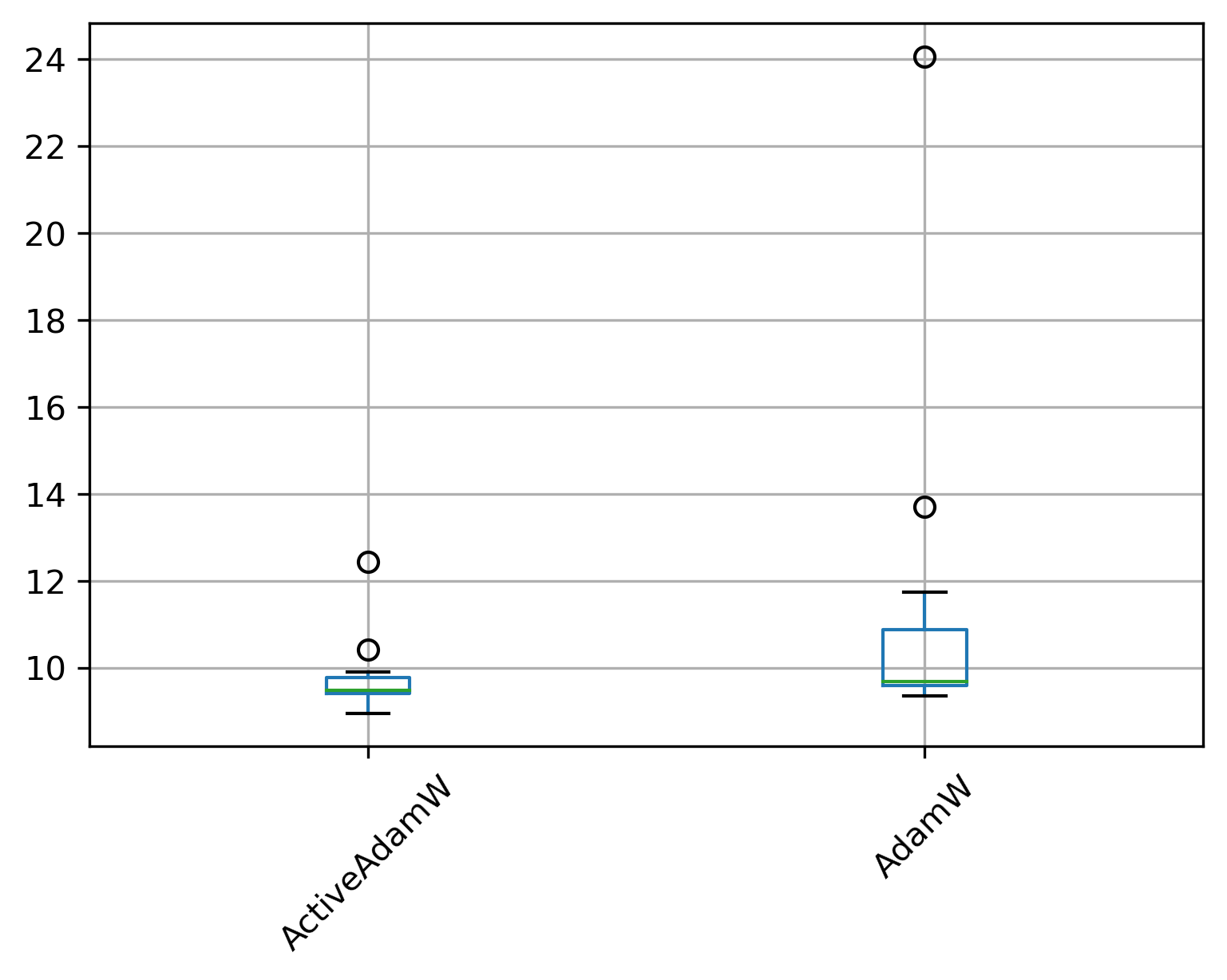}}
          \caption{Test-set PPL (lower is better)}
    \end{subfigure}%
    \hfill
      \begin{subfigure}[b]{.49\columnwidth}
        
        \centerline{\includegraphics[width=\columnwidth]{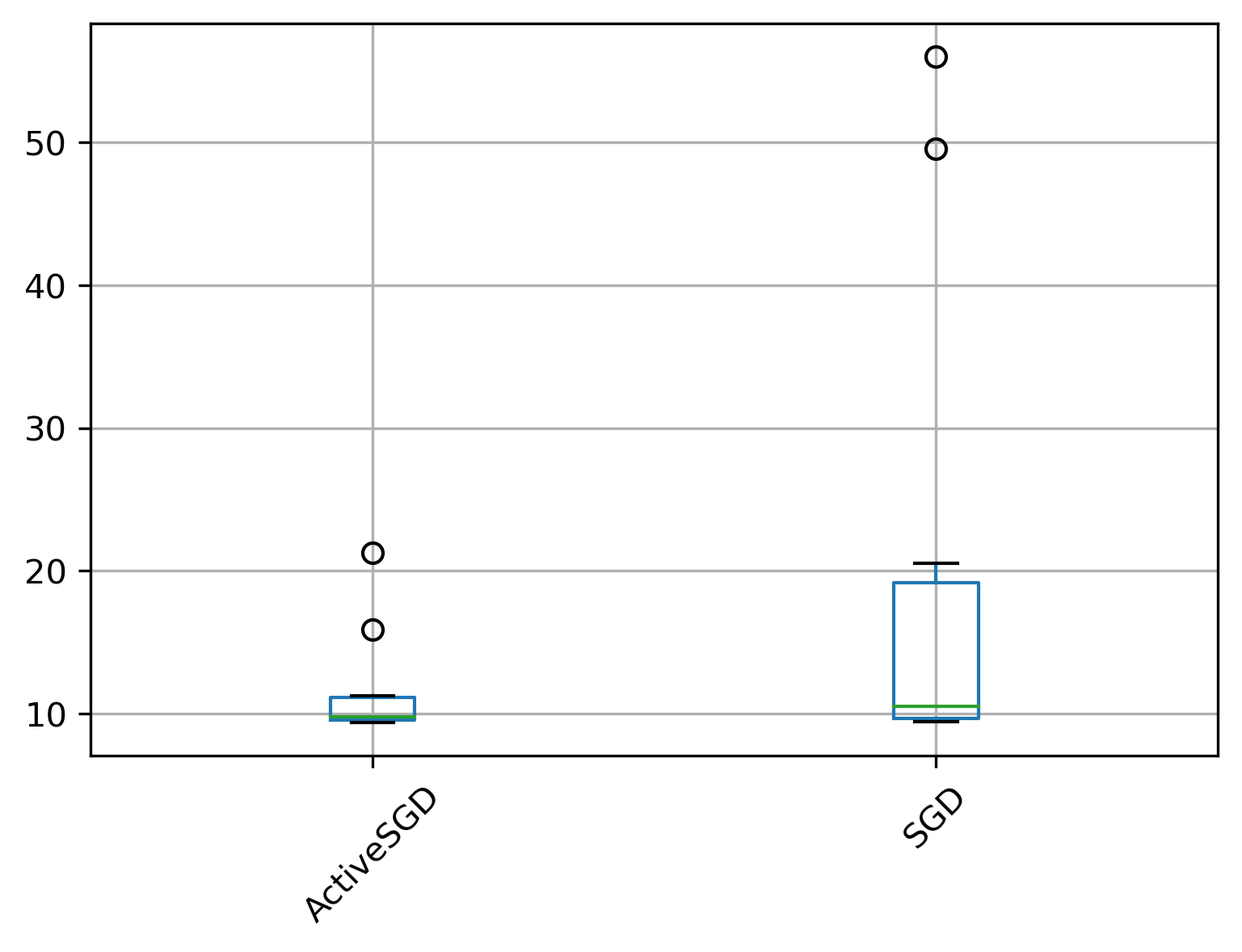}}
        \caption{Test-set PPL (lower is better).}
    \end{subfigure}
    \hfill
  
    \begin{subfigure}[b]{.49\columnwidth}
        
        \centerline{\includegraphics[width=\columnwidth]{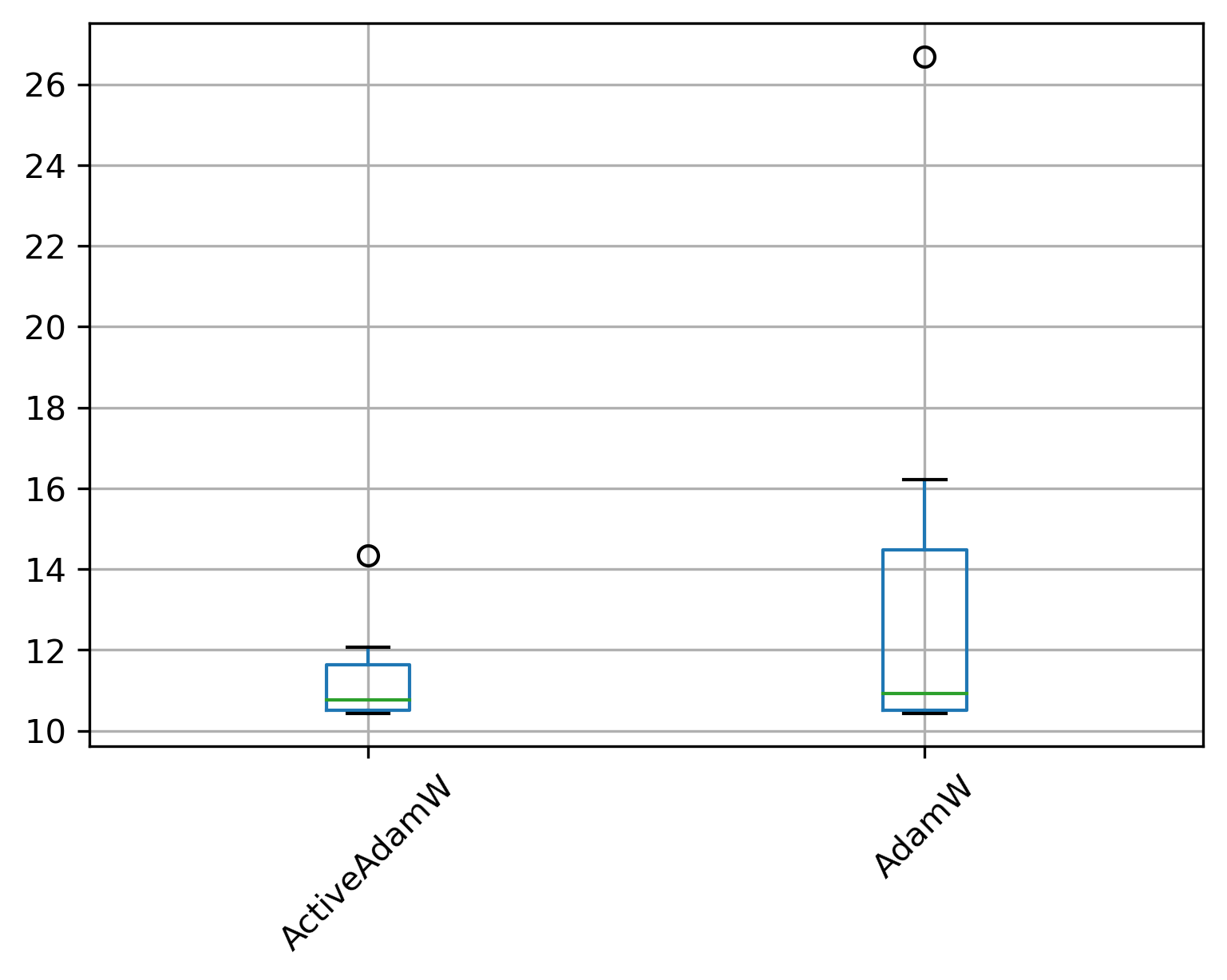}}
          \caption{Train-set PPL (lower is better)}
    \end{subfigure}
    \hfill
      \begin{subfigure}[b]{.49\columnwidth}
        
        \centerline{\includegraphics[width=\columnwidth]{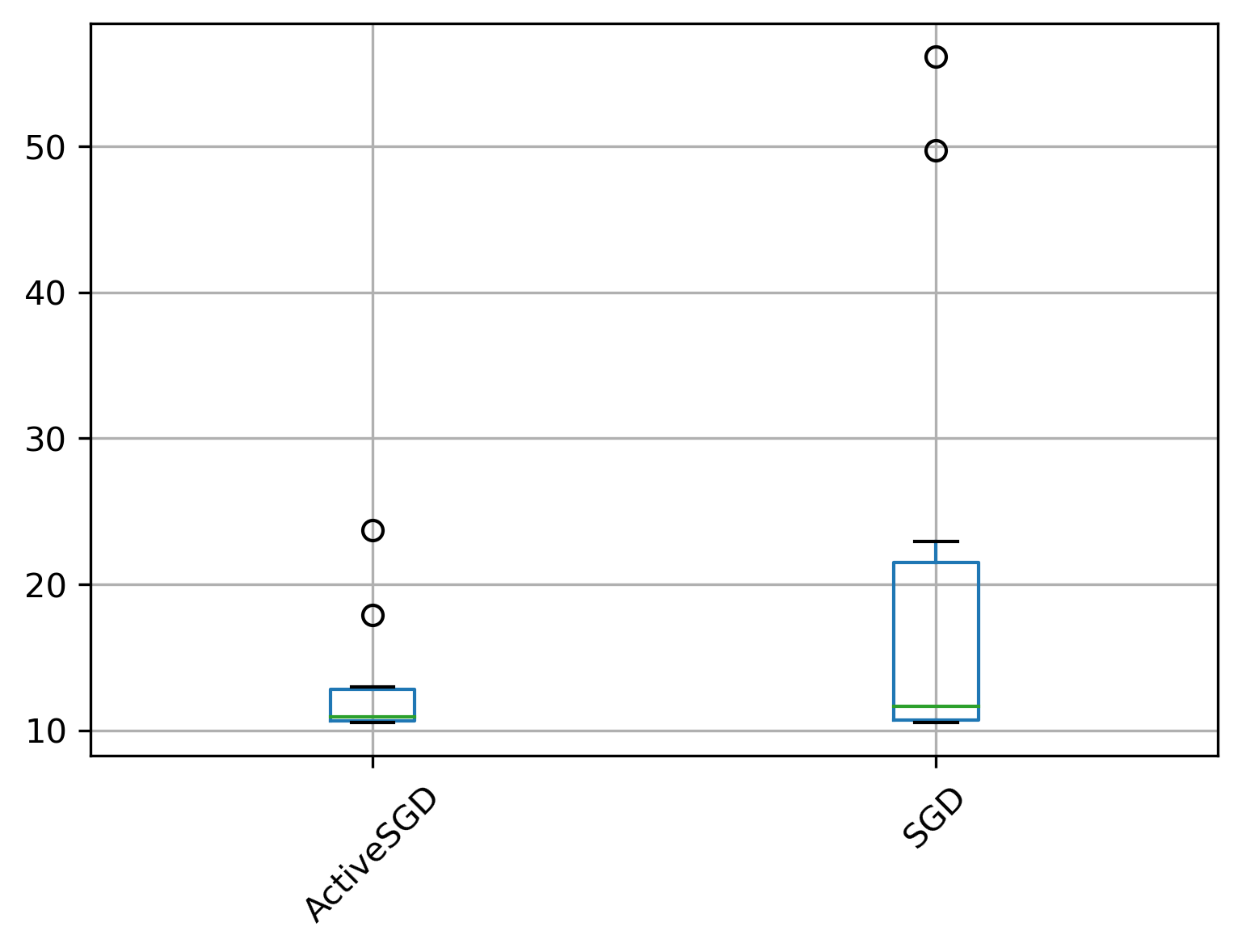}}
        \caption{Train-set PPL (lower is better).}
    \end{subfigure}
    
    \caption{WikiText-103 on mini-GPT-2 with various initial learning rates}
    \label{fig:wiki103full}
\end{center}
\vskip -0.2in
\end{figure}
  
\paragraph{PASCAL VOC on Faster-RCNN} 
We train PASCAL VOC on Faster-RCNN \cite{ren2015faster} with pretrained ResNet-50 backbone, following \cite{zhuang2020adabelief}.

\begin{figure}[ht]
    \vskip 0.2in
    \begin{center}
    \begin{subfigure}[t]{\columnwidth}
        
        \centerline{\includegraphics[width=\columnwidth]{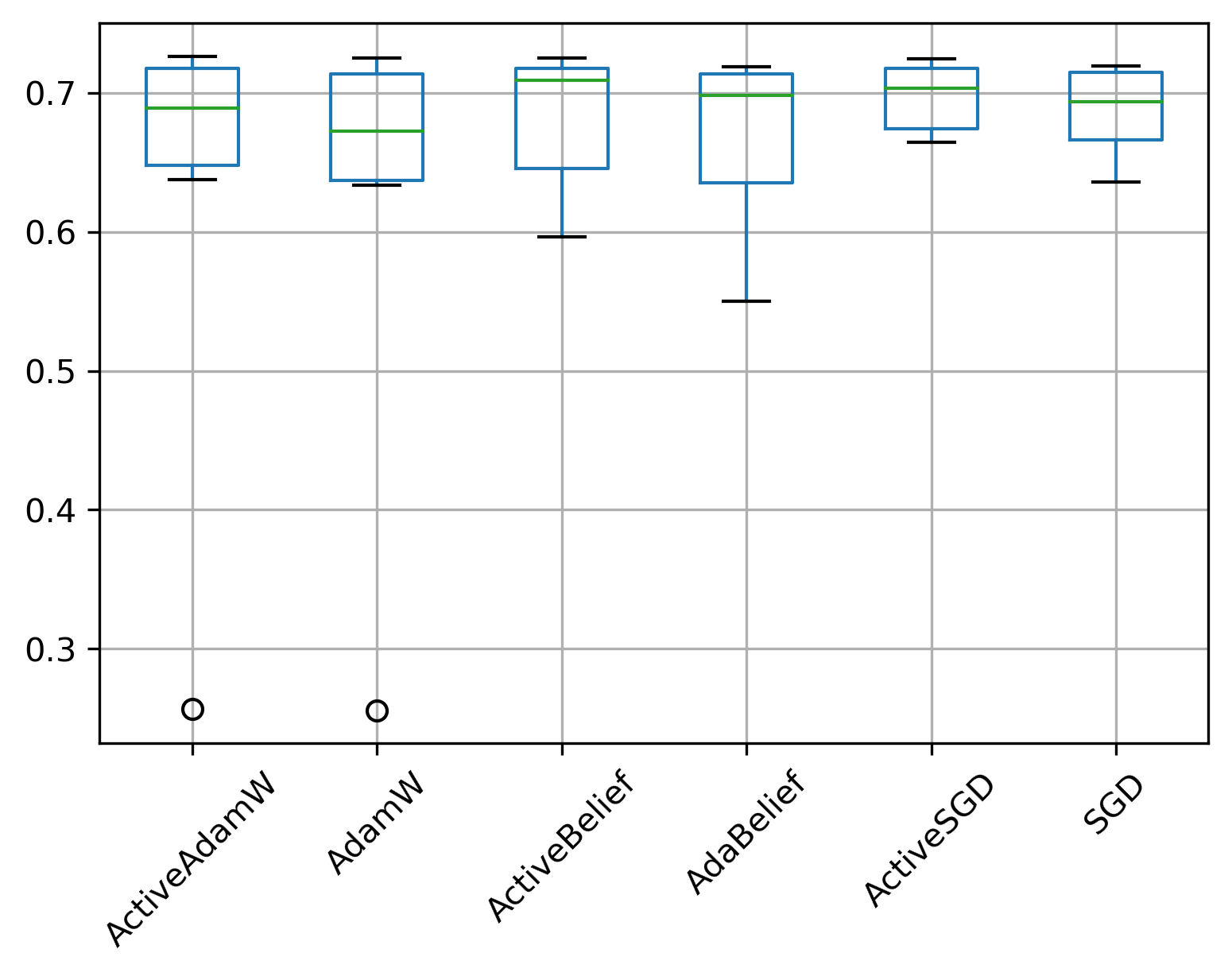}}
          \caption{Test-set mAP (higher is better)}
    \end{subfigure}
    \caption{PASCAL VOC on Faster-RCNN with various initial learning rates}
    \label{fig:pascal}
\end{center}
\vskip -0.2in
  \end{figure}

The results are shown in Figure \ref{fig:pascal}. Note that we could not replicate the mAP from \cite{zhuang2020adabelief}; we suspect the reason is their use of the MMDetection \cite{https://doi.org/10.48550/arxiv.1906.07155} framework, which does various extra image augmentation transforms.
  
\subsection{The choice of $\alpha_{low}$ and $\alpha_{high}$} \label{appendix:lrlowhigh}
We run a simulation to see the range of the learning rate after going through a series of $\alpha_{low}$ and $\alpha_{high}$ updates. We simulate a setting where the learning rate, $\alpha$, with 0.5 probability, is decreased by $\alpha_{low}$, otherwise, it is increased by $\alpha_{high}$. The simulation is run for $10^4$ epochs.

\begin{figure}[ht]
    \vskip 0.2in
    \begin{center}
  \begin{subfigure}[t]{\columnwidth}
  	
    \centerline{\includegraphics[width=\columnwidth]{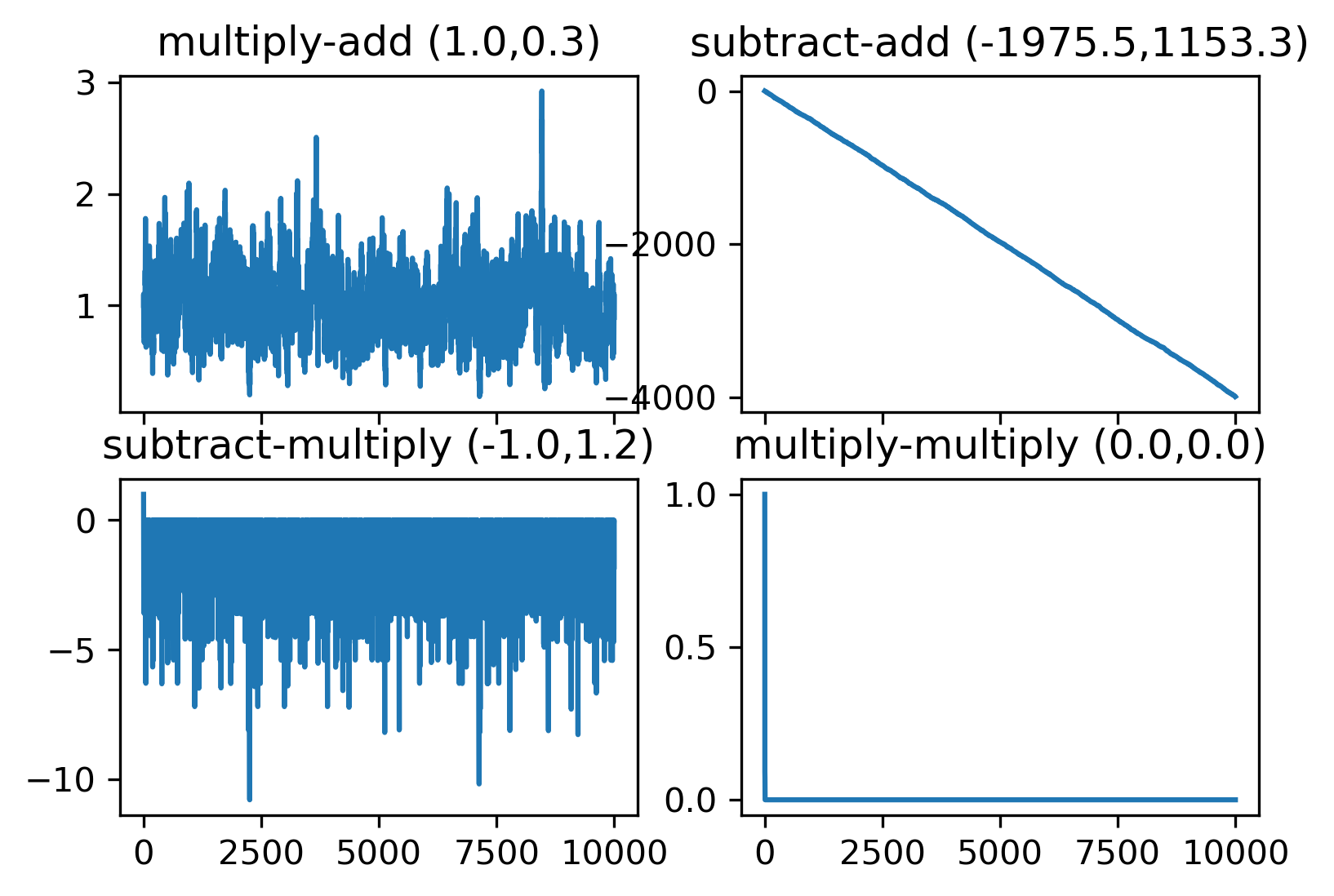}}
  \end{subfigure}
  
  \caption{learning rate (y-axis) plotted against training epochs (x-axis)}
  \label{fig:lrlowhigh}
\end{center}
\vskip -0.2in
\end{figure}

In Figure \ref{fig:lrlowhigh} we see that, consistent with \cite{tieleman2012lecture}, only multiplying by $\alpha_{low}$ and adding to $\alpha_{high}$ maintain a stable and positive learning rate ($\mu_{\alpha}=1.0$, $\sigma_{\alpha}=0.3$).
Other combinations lead to a negative learning rate (subtract-add and subtract-multiply) or shrink the learning rate to $0$ (multiply-multiply).

\subsection{Proofs}\label{proofs}
\paragraph{Theorem \ref{theorem:1}} In the local convex regime of a non-convex objective function, 
the cumulative gradient of parameter $\theta_i$ at epoch $e$ with no inner-loop updates, $\cums$, 
has the same sign as the cumulative gradient of parameter $\theta_i$ at epoch $e$ with inner-loop updates, $\cumh$, 
if the learning rate, $\alpha$, is smaller than $\alpha^*$ that causes the inner-loop to diverge.
\begin{equation}
\cums \cumh \geq 0
\end{equation}

\begin{proof} We prove this for the general case of the cost function $f$ for inner-loop updates of SGD. We wish to show that at time $K$, for a dataset of $K$ mini-batches, the cumulative gradient when there is no inner-loop updates to the parameters, $\cums$, has the same sign as the cumulative gradient when there is SGD updates at each iteration to the parameters, $\cumh$.
\newline
Suppose we are optimizing parameters $\theta$ using a convex objective function $f(\theta_t, batch_t)$, where $\theta_t$ and $batch_t$ are the parameter and the mini-batch at time-step $t$. For ease of notation, we will refer to the objective function evaluated for $batch_t$ as $f_t$. We will also refer to the gradients of the objective function w.r.t. the parameter $\theta_t$ for mini-batch $t$, as $g_t(\theta_t)$. When there are no inner-loop updates, we will refer to the parameter as $\theta_t^*$, which will equal the initial $\theta$, i.e., $\theta_0$, regardless of the value of $t$. When there are SGD updates in the inner-loop, we will refer to the parameter as $\widehat{\theta_t}$, which will be determined by the SGD algorithm \ref{thetahat}.
Formally,
\begin{equation} \label{theta0}
\begin{split}
\theta_t^* &= \theta_0
\end{split}
\end{equation}

\begin{equation} \label{thetahat}
\begin{split}
\widehat{\theta_t} &= \widehat{\theta_{t-1}} - \alpha g_t(\widehat{\theta_{t-1}}),
\end{split}
\end{equation}

where $\alpha$ is the constant, positive learning rate and  

\begin{equation} \label{g}
\begin{split}
g_t(\theta) = \frac{\partial{f_t(\theta)}}{\partial{\theta}}
\end{split}
\end{equation}

The no-update and SGD-updated cumulative gradients are defined as follows:
\begin{equation} \label{eqcus}
\begin{split}
\cums &= \sum_{t=1}^K g_t(\theta_{t-1}^*)  \\
\cumh &= \sum_{t=1}^K g_t(\widehat{\theta_{t-1}})
\end{split}
\end{equation}

Since \ref{thetahat} is recursive, we can unroll it from $\widehat{\theta_{t-1}}$ through $\widehat{\theta_{0}}$:

\begin{equation} \label{sgdunroll}
\begin{split}
\widehat{\theta_{t}} = \theta_{0} - \alpha \sum_{i=1}^t g_i(\widehat{\theta_{i-1}})
\end{split}
\end{equation}

In order to prove the theorem, we will use the following property of convex functions:

Supposing that f is a convex function, $a$ and $b$ two points in the domain of f, and $\frac{\partial{f(b)}}{\partial{b}}$ the derivative of f at point b, we have:

\begin{equation} \label{convex}
f(a) - f(b) \geq \frac{\partial{f(b)}}{\partial{b}} (a-b)
\end{equation}

We use \ref{convex} on $\widehat{\theta_K}$, i.e., the final parameter after the updates, and $\theta_0$, the initial parameter, for the objective function evaluated on batch $t$:

\begin{equation} \label{}
\begin{split}
f_t(\widehat{\theta_K}) - f_t(\theta_0) \geq \frac{\partial{f_t(\theta_0)}}{\partial{\theta_0}} (\widehat{\theta_K} - \theta_0)
\end{split}
\end{equation}

Replacing $\frac{\partial{f_t(\theta_0)}}{\partial{\theta_0}}$ with $g_t(\theta_0)$ according to \ref{g}:

\begin{equation} \label{}
\begin{split}
f_t(\widehat{\theta_K}) - f_t(\theta_0) \geq g_t(\theta_0) (\widehat{\theta_K} - \theta_0)
\end{split}
\end{equation}

Replacing $\widehat{\theta_K}-\theta_0$ with the unrolled version from \ref{sgdunroll}:

\begin{equation} \label{eqtmp1}
\begin{split}
f_t(\widehat{\theta_K}) - f_t(\theta_0) \geq - \alpha g_t(\theta_0) \sum_{i=1}^K g_i(\widehat{\theta_{i-1}})
\end{split}
\end{equation}

From the theorem's assumption that the optimization does not diverge, it follows that the objective function for the final parameter $\theta_k$ should have a lower value than the initial parameter $\theta_0$:

\begin{equation} \label{eqtmp2}
\begin{split}
f_t(\widehat{\theta_K}) - f_t(\theta_0) \leq 0
\end{split}
\end{equation}

Combining \ref{eqtmp1} and \ref{eqtmp2}, we have:

\begin{equation} \label{eqtmp3}
\begin{split}
0 \geq f_t(\widehat{\theta_K}) - f_t(\theta_0) \geq - \alpha g_t(\theta_0) \sum_{i=1}^K g_i(\widehat{\theta_{i-1}})\\
0 \leq f_t(\theta_0) -f_t(\widehat{\theta_K}) \leq \alpha g_t(\theta_0) \sum_{i=1}^K g_i(\widehat{\theta_{i-1}})
\end{split}
\end{equation}

From \ref{eqcus}, we replace $\sum_{i=1}^K g_i(\widehat{\theta_{i-1}})$ with $\cumh$:

\begin{equation} \label{eqtmp4}
\begin{split}
0 \leq f_t(\theta_0) - f_t(\theta_K) \leq \alpha g_t(\theta_0) \cumh
\end{split}
\end{equation}

Finally, we sum over all the mini-batches in the dataset, t=1...K:

\begin{equation} \label{eqtmp5}
\begin{split}
&0 \leq \sum_{t=1}^K[f_t(\theta_0) - f_t(\widehat{\theta_K})] \leq \sum_{t=1}^K[\alpha g_t(\theta_0) \cumh]\\
&0 \leq \sum_{t=1}^K[f_t(\theta_0) - f_t(\widehat{\theta_K})] \leq \alpha \cumh \sum_{t=1}^K g_t(\theta_0)
\end{split}
\end{equation}

From \ref{theta0} and \ref{eqcus}, $\sum_{t=1}^K g_t(\theta_0) = \cums$:

\begin{equation} \label{eqtmp6}
\begin{split}
0 \leq \sum_{t=1}^K[f_t(\theta_0) - f_t(\widehat{\theta_K})] \leq \alpha \cumh \cums
\end{split}
\end{equation}

Since $\alpha$ is positive:

\begin{equation} \label{eqtmp7}
\begin{split}
0 \leq \frac{\sum_{t=1}^K[f_t(\theta_0) - f_t(\widehat{\theta_K})]}{\alpha} \leq \cumh \cums \qed
\end{split}
\end{equation}

We see that $\cumh \cums$ is non-negative and found a lower bound $\frac{\sum_{t=1}^K[f_t(\theta_0) - f_t(\widehat{\theta_K})]}{\alpha}$ for it.
\end{proof}

\paragraph{Theorem \ref{theorem:2}}
Let us call the cost of vanilla SGD at epoch e, $f(\theta^S_e)$, and the cost of its corresponding ActiveLR implementation, $f(\theta^A_e)$. We define ActiveSGD's gradient, $g_{e+1}^A$, as $\frac{\partial{f(\theta_e^A)}}{\partial{\theta_e^A}}$, and vanilla SGD's gradient, $g_{e+1}^S$, as $\frac{\partial{f(\theta_e^S)}}{\partial{\theta_e^S}}$. $\alpha$ is the initial learning rate for ActiveLR SGD and also the constant learning rate for SGD. $\alpha_{High}$ is the higher learning rate ($\alpha_{High}>\alpha$) that ActiveLR uses when $g_e^A g_{e+1}^A >0$ and $\alpha_{Low}$ the lower learning rate ($\alpha_{Low}<\alpha$) that ActiveLR uses when $g_e^A g_{e+1}^A <0$. In the local convex regime of $f$, at any arbitrary epoch, $\mathit{e}$, the difference between the cost of using vanilla SGD and ActiveLR SGD at the next epoch, $\mathit{e+1}$, is 

\begin{equation}
\begin{cases}
f(\theta^S_{e+1}) - f(\theta^A_{e+1}) \geq g_{e+2}^A g_{e+1}^A (\alpha_{High}-\alpha),\\ \text{if } g_e^A g_{e+1}^A >0.\\

f(\theta^S_{e+1}) - f(\theta^A_{e+1}) \geq g_{e+2}^A g_{e+1}^A (\alpha_{Low}-\alpha),\\ \text{if } g_e^A g_{e+1}^A <0.
\end{cases}
\end{equation}
where the right hand side for both cases is non-negative.

\begin{proof}
At the start, the parameters and gradients are equal for ActiveLR and vanilla SGD: 
\begin{equation} \label{thetae}
\begin{split}
\theta_e^A &= \theta_e^S\\
g_{e+1}^A &= g_{e+1}^S
\end{split}
\end{equation}
$\theta_{e+1}^A$ and $\theta_{e+1}^S$ are as follows:

\begin{equation}
\theta_{e+1}^S = \theta_e^S - \alpha g_{e+1}^S\\
\end{equation}

Replacing $\theta_e^S$ with $\theta_e^A$ and $g_{e+1}^S$ with $g_{e+1}^A$,

\begin{equation} \label{thetae1s}
\theta_{e+1}^S = \theta_e^A - \alpha g_{e+1}^A
\end{equation}

For ActiveLR,
\begin{equation} \label{thetae1a}
\begin{split}
\begin{cases}
&\theta_{e+1}^A = \theta_e^A - \alpha_{High} g_{e+1}^A,\\ &\text{if } g_e^A g_{e+1}^A >0.\\
&\theta_{e+1}^A = \theta_e^A - \alpha_{Low} g_{e+1}^A,\\ &\text{if } g_e^A g_{e+1}^A <0.
\end{cases}
\end{split}
\end{equation}
For a differentiable convex function, f, at two points a and b in its domain, we have
\begin{equation} \label{convexineq}
f(a) \geq f(b) + \frac{\partial{f(b)}}{\partial{b}}(a-b).
\end{equation}

From \ref{convexineq}, for $\theta_{e+1}^A$ and $\theta_{e+1}^S$ we have

\begin{equation} \label{convexineqtheta}
f(\theta_{e+1}^S) \geq f(\theta_{e+1}^A) + \frac{\partial{f(\theta_{e+1}^A)}}{\partial{\theta_{e+1}^A}} (\theta_{e+1}^S-\theta_{e+1}^A)
\end{equation}

Replacing $\frac{\partial{f(\theta_{e+1}^A)}}{\partial{\theta_{e+1}^A}}$ with $g_{e+2}^A$

\begin{equation} \label{convexineqg}
f(\theta_{e+1}^S) - f(\theta_{e+1}^A) \geq g_{e+2}^A (\theta_{e+1}^S-\theta_{e+1}^A)
\end{equation}

Replacing from \ref{thetae1s} and \ref{thetae1a} into \ref{convexineqg}:

\begin{equation} \label{convexineqgrep}
\begin{split}
\begin{cases}
&f(\theta_{e+1}^S) - f(\theta_{e+1}^A) \geq g_{e+2}^A g_{e+1}^A (\alpha_{High}-\alpha),\\ &\text{if } g_{e}^A g_{e+1}^A>0.\\
&f(\theta_{e+1}^S) - f(\theta_{e+1}^A) \geq g_{e+2}^A g_{e+1}^A (\alpha_{Low}-\alpha),\\ &\text{if } g_{e}^A g_{e+1}^A<0.
\end{cases}
\end{split}
\end{equation}

We can divide the training process into two segments, $e_0$ to $e_{switch}$ when the gradients change after each epoch, and $e_{switch+1}$ to  the end of training, $e_T$, when the gradients do not change. Note that the opposite can be also be true, i.e., $e_0$ to $e_{switch}$ is when the gradients do not change after each epoch, and $e_{switch+1}$ to the end of training, $e_T$, when the gradients change after each epoch. Here we treat the former. After the proof of the former, the latter can be proved trivially.

While we are in the segment $e_0$ to $e_{switch}$, since the gradients change sign after each epoch, $g_{e+2}^A g_{e+1}^A<0$ and $g_{e}^A g_{e+1}^A<0$. Therefore, the second inequality of \ref{convexineqgrep} holds. On the right hand side, we have $\alpha_{Low}-\alpha<0$ from the theorem's assumption. As a result, the right hand side is non-negative, and $f(\theta_{e+1}^S) - f(\theta_{e+1}^A)$ is greater than or equal to a non-negative number.

When we are in the segment $e_{switch+1}$ to $e_{T}$, since the gradients do not change sign after each epoch, $g_{e+2}^A g_{e+1}^A>0$ and $g_{e}^A g_{e+1}^A>0$. Therefore, the first inequality of \ref{convexineqgrep} holds. On the right hand side, we have $\alpha_{High}-\alpha>0$ from the theorem's assumption. As a result, the right hand side is non-negative, and $f(\theta_{e+1}^S) - f(\theta_{e+1}^A)$ is greater than or equal to a non-negative number.
\end{proof}

\subsection{Interactive webapp}
We created an interactive webapp where the readers can compare the performance of ActiveAdam and Adam in minimizing 
a saddle and an MSE loss function, with the ability to change hyper-parameter values, such as initial learning rate, 
initial weight, and number of epochs.
The webapp can be accessed through the link below:\\
\url{https://active-lr.herokuapp.com}

\section{Notes on the appended codes}
The training scripts are written for SLURM workload manager since our experiments were run on HPCs with SLURM.

\subsection{License}\label{license}
The licenses for the assets used in this paper are the following:
\begin{itemize}
\item AdamW: This work is licensed under a BSD 3-Clause "New" or "Revised" License
\item RAdam: This work is licensed under a Apache License 2.0
\item AdaBelief: This work is licensed under a BSD 2-Clause "Simplified" License
\item CIFAR-10 and -100: This work is licensed under a New BSD License
\item WikiText-2 and -103: This work is licensed under a Creative Commons Attribution-ShareAlike License
\item ImageNet: This work is licensed under a custom (research, non-commercial) license
\item GPT-2: This work is licensed under a Modified MIT License
\end{itemize}


\end{document}